\documentclass{article}

    \PassOptionsToPackage{numbers, compress}{natbib}

\usepackage{amsmath} 
\usepackage{natbib}  

\usepackage[final]{neurips_2025}

\usepackage[utf8]{inputenc} 
\usepackage[T1]{fontenc}    
\usepackage{hyperref}       
\usepackage{url}            
\usepackage{booktabs}       
\usepackage{amsfonts}       
\usepackage{nicefrac}       
\usepackage{microtype}      
\usepackage{xcolor}         
\usepackage{amsmath}
\definecolor{OliveGreen}{RGB}{107,142,35}
\usepackage{pifont}
\usepackage[svgnames,dvipsnames]{xcolor}
\definecolor{nipsblue}{rgb}{0.21,0.49,0.74}

\usepackage{graphicx}

\title{%
  \includegraphics[height=1cm]{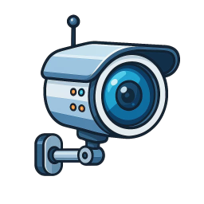} MoniTor: Exploiting Large
  Language Models with Instruction for Online Video Anomaly Detection
}

\begin{document}
%

\author{
  Shengtian Yang\textsuperscript{*},
  Yue Feng\textsuperscript{*},
  Yingshi Liu\textsuperscript{},
  Jingrou Zhang\textsuperscript{},
  Jie Qin\textsuperscript{\dag} \\
  College of Artificial Intelligence, Nanjing University of Aeronautics
and Astronautics \\
Key Laboratory of Brain-Machine Intelligence
Technology, Ministry of Education, China
}

\renewcommand{\thefootnote}{*} 
\footnotetext[1]{Equal contribution.}
\renewcommand{\thefootnote}{\dag} 
\footnotetext[2]{Corresponding author.}


\maketitle

\begin{abstract}
Video Anomaly Detection (VAD) aims to locate unusual activities or behaviors within videos. Recently, offline VAD has garnered substantial research attention, which has been invigorated by the progress in large language models (LLMs) and vision-language models (VLMs), offering the potential for a more nuanced understanding of anomalies. 
However, online VAD has seldom received attention due to real-time constraints and computational intensity. 
In this paper, we introduce a novel \textbf{M}emory-based online scoring queue scheme for \textbf{T}raining-free VAD (MoniTor), to address the inherent complexities in online VAD. 
Specifically, MoniTor applies a streaming input to VLMs, leveraging the capabilities of pre-trained large-scale models. 
To capture temporal dependencies more effectively, we incorporate a novel prediction mechanism inspired by Long Short-Term Memory (LSTM) networks. This ensures the model can effectively model past states and leverage previous predictions to identify anomalous behaviors. Thereby, it better understands the current frame. 
Moreover, we design a scoring queue and an anomaly prior to dynamically store recent scores and cover all anomalies in the monitoring scenario, providing guidance for LLMs to distinguish between normal and abnormal behaviors over time.
We evaluate MoniTor on two large datasets (i.e., UCF-Crime and XD-Violence) containing various surveillance and real-world scenarios. 
The results demonstrate that MoniTor outperforms state-of-the-art methods and is competitive with weakly supervised methods without training. Code is available at \url{https://github.com/YsTvT/MoniTor}. 
\end{abstract}

\begin{figure}
    \centering
    \includegraphics[width=0.9\linewidth]{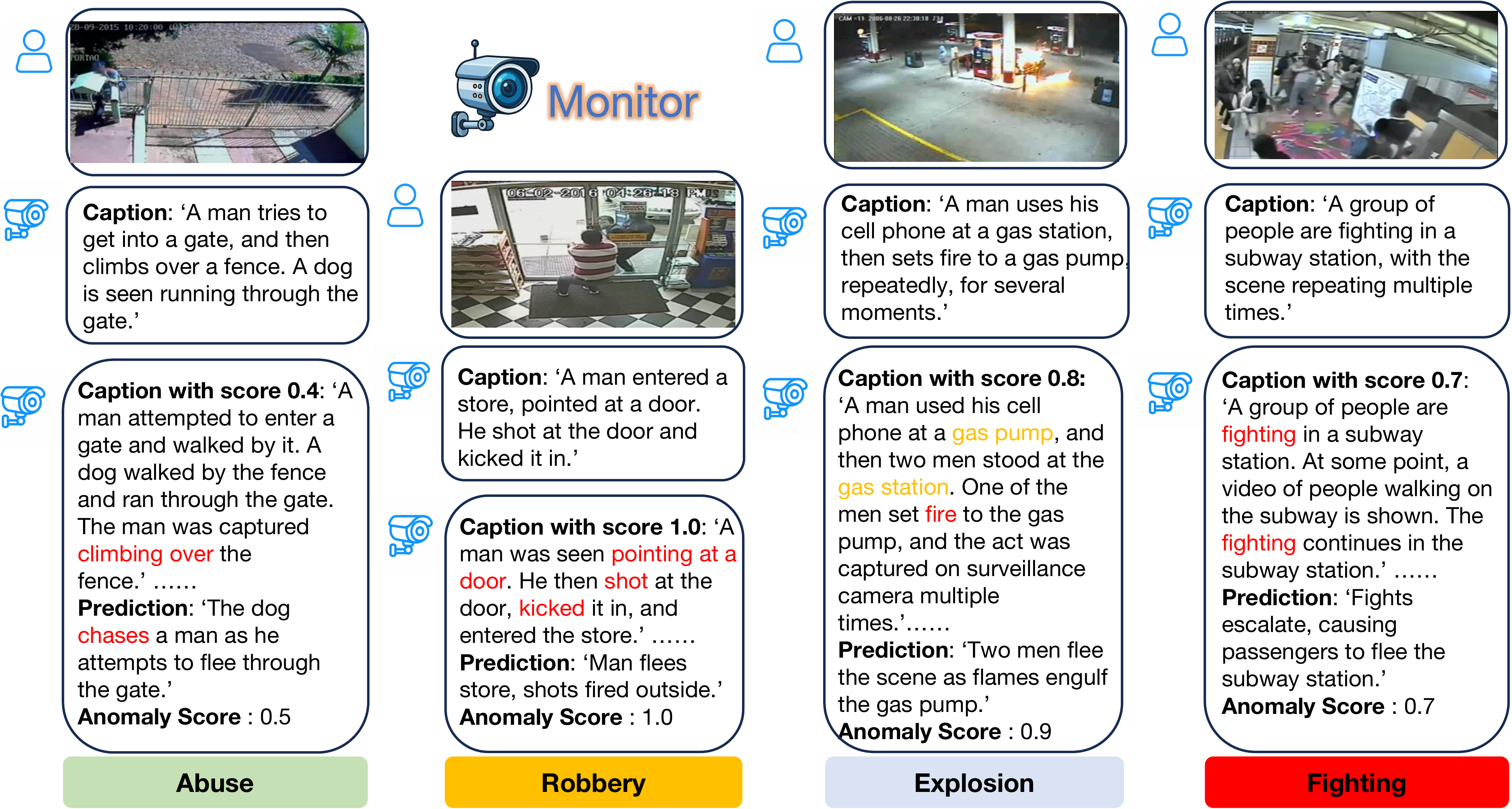}
    \caption{Illustration of MoniTor which detects abnormal events across multiple surveillance perspectives. MoniTor identifies critical security incidents: Abuse, Robbery, Explosion, Fighting and so on.}
    \label{fig:intro}

\end{figure}

\section{Introduction}
\label{sec:intro}
Video Anomaly Detection (VAD) aims to locate abnormal activities or behaviors in videos, which is crucial for video understanding applications~\cite{ji2023multispectral,ge2025implicitlocationcaptionalignmentcomplementary,RefineTAD,10.1145/3581783.3612506}.
However, existing VAD methods~\cite{9523042, wu2022self, tian2021weakly, 2022arXiv221115098C} are mostly in an offline fashion, ignoring the demands for real-time monitoring and real-world applications, which also play an important role in many real-life scenarios, such as intelligent surveillance~\cite{survellience1, survellience2,Jia_Quan_Feng_Chen_Qin_2025,su2024maca,10817642}, autonomous driving~\cite{autodriving}, etc. 

Compared to offline VAD, anomaly detection can be further complicated in scenarios where data arrive in a streaming/online manner, especially when it is required to identify anomalies as they occur.
The difficulties lie in, moreover, the inherent characteristics of online anomalies, because they are discontinuous and occur infrequently in real scenarios, which results in a scarcity of extensive and diverse anomaly data for training. 
Moreover, the high complexity of human behaviors (i.e., encompassing a vast array of both normal and abnormal actions) poses obstacles to the generalizability of VAD models in real-world settings. Current datasets fail to comprehensively capture the diversity of human behaviors. This significantly limits the VAD model’s generalization ability across different domains and scenarios.
For example, Karim \textit{et al.}~\cite{karim2024real} introduced REWARD, a weakly-supervised framework for real-time anomaly detection. Although trained as an end-to-end video model, REWARD struggles with dynamic camera angles and complex scenes due to limited training data, which limits its applicability across diverse scenarios.
Recent VAD solutions have also been devoted to tackling these challenges with pretrained large-scale models.
Zanella \textit{et al.}~\cite{LAVAD} proposed LAVAD, a training-free VAD approach utilizing Large Language Models (LLMs) to score potential anomalies directly from text, thus bypassing data collection and annotation. 
However, LAVAD is limited to offline VAD, as applying LLMs to online VAD faces additional challenges. 
Capturing historical information for anomaly scoring may lead to model misinterpretation when anomalous memories are encountered in normal videos.
In addition, LLMs' reliance on explicit instructions impedes their ability to genuinely identify anomalous events.

In this paper, we propose a novel \textbf{M}emory-based online scoring queue scheme for \textbf{T}raining-free VAD, namely \textbf{MoniTor}, to solve the above challenges. 
As shown in Fig.~\ref{fig:intro}, our MoniTor can precisely and efficiently identify various abnormal events. Firstly, we introduce a hierarchical dual-memory architecture through Dynamic Memory Gating Module that systematically addresses temporal discontinuity inherent in online anomalies. This architecture integrates a long-term episodic memory module with adaptive forgetting mechanisms and a short-term working memory encoding fine-grained spatiotemporal patterns. Through this dual-memory design, we effectively tackle the challenge of discontinuous and infrequent anomaly occurrences in real scenarios. Secondly, we formulate a principled anomaly scoring protocol via Standard Scoring Queue that incorporates a novel queuing mechanism for sequential anomaly descriptor propagation. This protocol leverages a knowledge-enhanced anomaly prior derived from encyclopedic sources. Such design significantly expands the model's generalization capacity across diverse anomalous events, addressing the obstacles posed by both the high complexity of human behaviors and the limitations of available datasets. Thirdly, we propose a predictive scoring framework  in Behavior Prediction and Dynamic Analysis component that exploits temporal causality in streaming video. This framework establishes a feedback loop between expectation and reality, improving detection sensitivity for emergent anomalies despite their stochastic and infrequent manifestation. Consequently, our approach effectively mitigates the scarcity of extensive and diverse anomaly data for training. Moreover, we conduct rigorous experimental validation on challenging benchmark datasets, i.e., UCF-Crime~\cite{2018arXiv180104264S} and XD-Violence~\cite{wu2020not}. Our comprehensive analysis demonstrates that MoniTor significantly outperforms state-of-the-art online unsupervised approaches and offline training-free methods across multiple evaluation metrics. These results empirically validate that our framework effectively captures temporal context and facilitates robust anomaly comprehension in LLMs, overcoming the significant restrictions on VAD models' effectiveness beyond specific datasets.

In summary, our contributions are four-fold:
\begin{itemize}
    \item We introduce MoniTor, which applies Large Language Models (LLMs) for online VAD. Our MoniTor facilitates real-time monitoring through streaming video inputs, with the notable capability of generating anomaly scores at 0.6-second intervals while maintaining a 5-second end-to-end processing latency.
    \item We integrate the Long Short-Term Memory (LSTM) networks with LLMs to effectively encode historical sequence information, which enhances the performance of online VAD and makes the identification of anomalous event boundaries more precisely.
    \item We propose an innovative scoring queue mechanism to mitigate the challenges associated with instruction dependency within LLMs. Furthermore, we introduce an anomaly prior, which is instrumental in training LLMs to effectively discern anomalous events.
    \item Extensive experiments demonstrate that our proposed MoniTor achieves superior performance compared to unsupervised approaches and surpasses training-free offline methods.
\end{itemize}

\section{Related work}

\textbf{Online VAD.}
In general, VAD is as an out-of-distribution detection problem and uses training data of different supervision levels to learn normal distribution, including full supervision (\textit{i.e.}, frame-level supervision of normal video and abnormal video)
~\cite{Bai_2019_CVPR_Workshops,Wang2019AnomalyCI,2021PatRe.11407865D,9523042,0Memory,8851288}, weak supervision (\textit{i.e.}, video-level monitoring of normal video and abnormal video)~\cite{2022arXiv221205136J,Li2022SelfTrainingML,2018arXiv180104264S}, one-class (\textit{i.e.}, only normal video)~\cite{liu2021hybrid,lv2021learning,9157029} and unsupervised (\textit{i.e.}, unlabeled video)
~\cite{2023arXiv230701533O,zaheer2022generative,5995524}. 
Video anomaly detection is categorized into online and offline fashion in the area of computer vision~\cite{2023arXiv230807050J}. 
Most of the existing work on offline VAD has made great breakthroughs. 
However, in real life, to avoid crime, we need to detect anomalies in the video in a timely manner. 
In the early research of online VAD, Chaker \textit{et al.}~\cite{chaker2017social} constructed a spatio-temporal cuboid using a window-based method to achieve online anomaly detection and localization. Luo \textit{et al.}~\cite{8019325} and Wang \textit{et al.}~\cite{8451070} encoded motion and appearance with LSTM Auto-Encoder. 
Recently, inspired by the dense video captioning streaming model which does not require access to all input frames proposed by Zhou \textit{et al.}~\cite{2024arXiv240401297Z}, Rossi \textit{et al.}~\cite{0Memory} proposed MOVAD equipped with two main components: a Short-Term Memory Module (STMM) and a Long-Term Memory Module (LTMM) to process past and current frames for online VAD tasks.
However, existing online VAD models exhibit certain limitations. Some fail to effectively capture historical information, while others may produce scores that are skewed by misleading historical data.
\\
\textbf{LLM-based VAD.} 
Recently, with the emergence of powerful LLMs such as GPT~\cite{achiam2023gpt,brown2020language,radford2019language} and Llama~\cite{touvron2023llama,touvron2023llama2}, several notable VAD approaches have leveraged these models. Kim \textit{et al.}\cite{s23146256} employed ChatGPT for textual descriptors coupled with VLM-based anomaly detection, while Zanella \textit{et al.}\cite{LAVAD} pioneered a training-free paradigm using Llama-2~\cite{touvron2023llama2} to generate anomaly scores from BLIP-2~\cite{li2023blip} frame descriptions.
However, current LLM-based VAD approaches exhibit fundamental limitations: they lack robust contextual reasoning capabilities for temporal reasoning in video sequences and demonstrate high sensitivity to prompt engineering, resulting in inconsistent performance when instructions are ambiguous. Despite these constraints, LLMs offer crucial advantages over traditional approaches that require exhaustive domain-specific training. Specifically, LLMs enable effective domain adaptation without data collection overhead or retraining, making them particularly suitable for diverse, cross-domain deployment scenarios.
Our approach addresses these limitations through two key technical innovations: (1) an LSTM-based forgetting gate mechanism that selectively preserves temporal context while eliminating irrelevant information, and (2) a novel scoring queue architecture that provides structured guidance to the LLM, substantially enhancing its decision-making precision in dynamic environments. Consequently, we present MoniTor, the first online training-free VAD framework that effectively leverages LLMs for real-time anomaly detection with robust temporal reasoning capabilities.

\section{Method}

\begin{figure*}[h]
    \centering
    \includegraphics[width=0.95\linewidth]{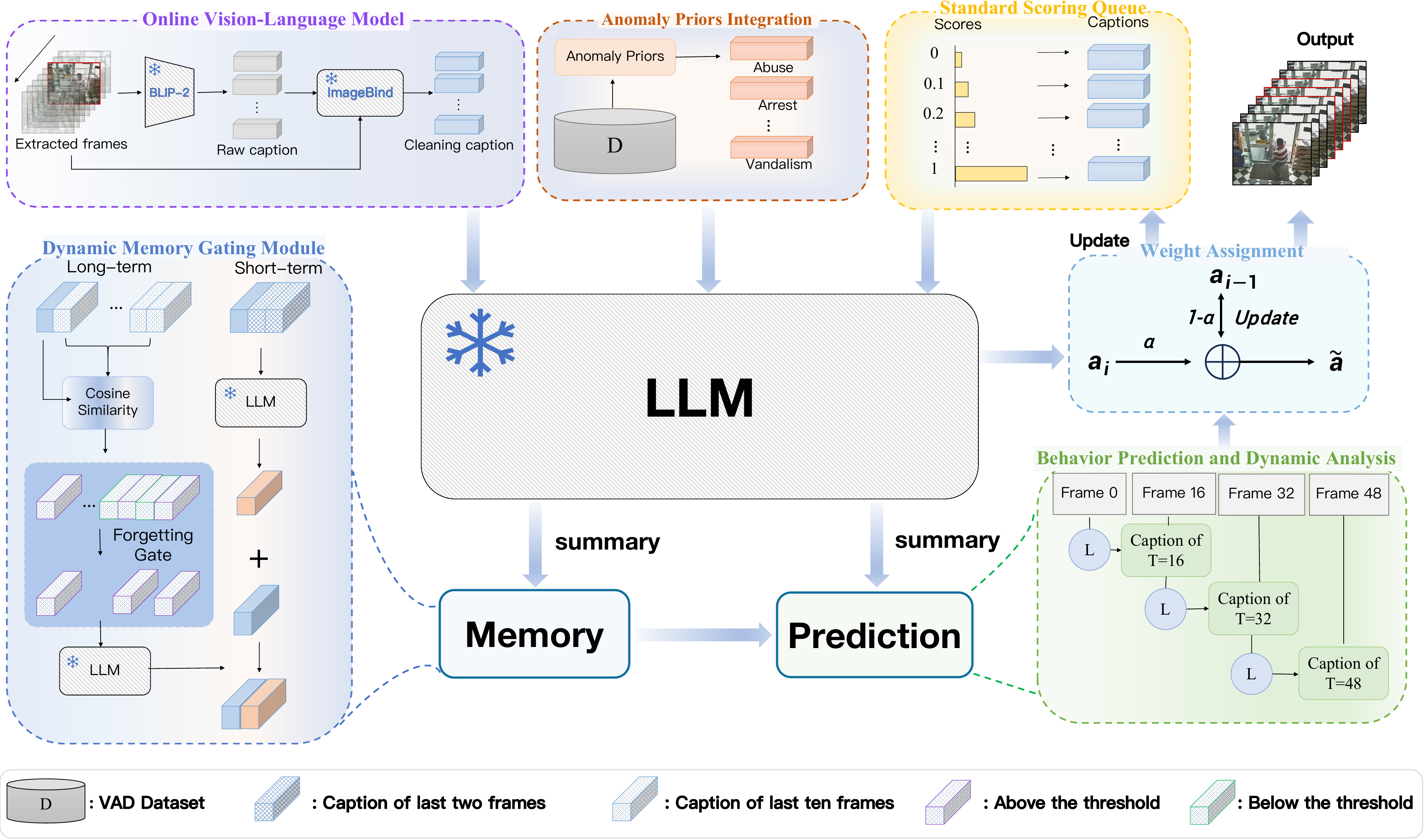}
    \caption{The architecture of our MoniTor: (1) Online Vision-Language Model (Sec.~\ref{me:ovlm}) is used to get each frame's textual summary. 
    (2) Anomaly Priors Integration (Sec.~\ref{me:api}) is used to serve as a form of ``knowledge injection''. 
    (3) Dynamic Memory Gating Module (Sec.~\ref{me:dmgm}) is used to capture historical information while preventing the large model from being misled by historical memory. 
    (4) Behavior Prediction and Dynamic Analysis (Sec.~\ref{me::bpada}) leverages frame-to-frame predictive cues to facilitate robust anomaly detection through comparative analysis of temporal discrepancies.
    (5) Standard Scoring Queue (Sec.~\ref{me:ssq}) is used to guide the large model on how to identify and understand anomalies. 
    (6) Score Optimization and Weight Assignment (Sec.~\ref{me:wa}) adjusts LLMs' scoring results based on different context, better aiding LLMs in distinguishing abnormal behaviors.}
    \label{fig2}

\end{figure*}
\textbf{Overview.}
The overall framework of our method is shown in Fig.~\ref{fig2}. Specifically, after a frame is extracted from the untrimmed video, it is fed into the Online Visual-Language Model to generate a textual summary. 
Then, the Anomaly Priors Integration is employed to guide the LLM to better understand the concept of anomalous events. To fully integrate historical information, the Dynamic Memory Gating Module respectively summarizes captions of long-term and short-term historical frames and pass them to the LLM. Meanwhile, the Behavior Prediction and Dynamic Analysis Module is applied to guide the LLM in generating a prediction for the caption of the next frame, which is passed to LLM when processing the next frame. 
Moreover, the Standard Scoring Queue is employed to store the corresponding historical frames for each score, and the anomaly score predicted by the LLM is used for updating the scoring queue.

\subsection{Online Vision-Language Model}~\label{me:ovlm}
Online Vision-Language Model is proposed to transform video frames online into their corresponding textual descriptions as LAVAD~\cite{LAVAD}. In this module, we first use five BLIP-2 models to generate five raw captions $R_i = \{{R_{i1}}, R_{i2}, R_{i3} ,R_{i4} ,R_{i5}\}$ for the current frame $\mathbf{I}_i$. However, the raw captions may be noisy. To mitigate this problem, we make full use of historical information. For each raw caption $A_j$ in $A = R_i \cup R_{i-1} \cup R_{i-2} \cup R_{i-3} \cup R_{i-4} \cup R_{i-5}$, we compute the cosine similarity $X_j$ between its text feature and the image feature of the frame:$
X_j = \langle \mathcal{E}_I(\mathbf{I}_i) \cdot \mathcal{E}_T(A_j)\rangle$,
where $\langle\cdot,\cdot\rangle$ is the cosine similarity, $\mathcal{E}_I$ is the image encoder of ImageBind, and $\mathcal{E}_T$ is the textual encoder of ImageBind. Afterthat, we sort all raw captions $A_j$ in $A$ by the cosine similarity $X_j$ and select the top 10 as cleaning captions $C_i = \{A_1, A_2, \ldots, A_{10}\}$. Finally, we send the cleaning captions $C_i$ into GLM-4-Flash to get the summary: $S_i = \Phi_\mathtt{GLM}(\texttt{P}_S \circ C_i)$, 
where prompt $\texttt{P}_S$ is formed as ``\textit{Please summarize what happened in few sentences, based on the following temporal description of a scene.}" $\circ$ is text concatenation. $\Phi_\mathtt{GLM}$ refers to generating summary through GLM-4-Flash.

\subsection{Anomaly Priors Integration}~\label{me:api}
In this module, since UCF-Crime and XD-Violence contain 13 and 6 categories of anomalies in surveillance scenarios, respectively, they cover a wide range of possible offences. We intend to include anomaly priors $P_A$ in context prompt to guide LLMs in recognizing anomalies and paying attention to them. We guide LLMs by adding the definitions of these anomalies from Wikipedia to the context prompt and giving examples as appropriate.

\subsection{Dynamic Memory Gating Module}~\label{me:dmgm}
This module is based on the LSTM architecture, capturing both long-term memory (\(M_l\)) and short-term memory (\(M_s\)). A forgetting gate is used to ensure the model accurately represents the input removing noise. Long-term memory (\(M_l\)) maintains a summary of text descriptions from frames over a 10-frame window, denoted as \(S_i = \{S_{i-10}, S_{i-9}, \dots, S_{i-1}\}\). These frames are filtered through the forgetting gate (\(F\)), which evaluates the similarity between the current and past frames. Frames with a similarity above a threshold $\theta$ are retained for summarization.
The long-term memory is updated as:

\begin{equation}
\label{eq:forget}
M_l = \Phi_\mathtt{GLM}(D_l), 
\end{equation}
\begin{equation}
D_l =
\begin{cases}
    D_1 + S_{i-j}, & \text{if}~d_{i-j} > \theta \\
    D_1, & \text{if}~d_{i-j} \leq \theta
\end{cases}
\quad j = 1, 2, \dots, 10
\end{equation}
\begin{equation}
\label{eq:forget-gate}
d_{i-j} = \langle \mathcal{E}_T(S_i), \mathcal{E}_T(S_{i-j}) \rangle,
\end{equation}
where \(\langle\cdot,\cdot\rangle\) represents cosine similarity, and the textual encoder \(\mathcal{E}_T : \mathcal{T} \rightarrow \mathcal{Z}\) maps text to vector space representations. The value \(d_{i-j}\) is the cosine similarity between frames \(S_i\) and \(S_{i-j}\), and \(D_l\) is the long-term memory storage filtered by the forgetting gate. \(j\) serves as the index for traversing from the current frame back to the previous 10 frames.


In contrast, short-term memory (\(M_s\)) summarizes the most recent two frames, \(S_{i-1}\) and \(S_{i-2}\), providing a more immediate representation of recent context. The short-term memory is updated by $M_s = \Phi_\mathtt{GLM}(D_s)$,
where \(D_s\) contains the text descriptions from the two previous frames.

\subsection{Behavior Prediction and Dynamic Analysis}~\label{me::bpada}
In this section, we focus on enhancing the model's ability to predict behavior and perform dynamic analysis by leveraging summarized information from both long-term and short-term memory. An LSTM-based architecture is utilized due to its effectiveness in analyzing sequential data, making it particularly suited for behavior prediction in video sequences. This approach also plays a critical role in the model's scoring phase. The prediction is obtained by:
$P_2 = \Phi_\mathtt{GLM}(P_{p} \circ S_{i} \circ P_{pf})$,
where \( S_{i} \) represents the summary of the current frame, and \( P_2 \) is the prediction for the next frame based on the current frame. The prediction step occurs within the scoring phase of the prior step. The component \( P_p \) is designed to prompt, ``\textit{If you are a law enforcement agency, predict what might happen next in this scene, taking into account possible suspicious activities or behaviors such as abuse, arrests, arson, assault, burglary, disorderly conduct, explosions, fights, robbery, shootings, theft, or vandalism. Provide a concise prediction based on the current context.}'' \( P_{pf} \) is structured to prompt,``\textit{Please predict concisely the behavior or event likely to occur next in the scene, avoiding any additional explanations.}''
This approach allows for a precise and targeted assessment of potential behaviors in dynamic video sequences, optimizing the model’s scoring and analytical capabilities.

\subsection{Standard Scoring Queue}~\label{me:ssq}
In this module, we implement a dynamic scoring system for LLMs, which helps guide the model to generate high-quality outputs based on predefined evaluation criteria. To achieve this, we maintain a scoring queue $Q = \{Q_{0}, Q_{0.1}, \ldots, Q_{1}\}$, where each element $Q_i$ represents the most recent caption that received a score of $i$.  The scoring queue serves as a repository of these anomaly assessments, enabling real-time updates and comparisons. It is updated as follows: $Q_{a_{i-1}} = S_{i-1}$, where \( a_{i-1} \) represents the anomaly score  of the \( i-1 \) frame, and \( S_{i-1} \) denotes the text description of the \( i-1 \) frame. The equation indicates that the summary of the \( i-1 \) frame \( S_{i-1} \) is stored at the corresponding position \( Q_{a_{i-1}} \) in the queue based on its anomaly score \( a_{i-1} \), which is used to record and track the anomaly detection result at that specific time and provide LLMs with guidance on scoring.

\subsection{Score Optimization and Weight Assignment}~\label{me:wa}
In our approach, we implement a dynamic weight assignment strategy to adaptively balance the importance of the current frame's score with the score of the previous frame. This mechanism enables the model to respond to changes in the video sequence while preserving continuity based on prior frames. By doing so, the model can gradually adjust to new information in each frame without abruptly discarding the historical context provided by earlier frames. 

The weight assignment process is structured as follows: for each frame in a batch, the score is computed by combining the current frame's score with the score of the previous frame, ensuring a weighted contribution from both. This balance is controlled by a parameter \( \alpha \), which determines the proportion of influence from the current and previous frames.
The weighted score is defined by:
$\tilde{a}_i = \alpha \times a_i + (1-\alpha) \times a_{i-1}$,
here, \( \tilde{a}_i \) represents the adjusted score for the current frame \( i \), \( a_i \) is the raw score of the current frame, and \( a_{i-1} \) is the score of the previous frame. The parameter \( \alpha \) (where \( 0 \leq \alpha \leq 1 \)) controls the weighting between the two scores, allowing for flexible adaptation to dynamic changes in the video sequence while still considering the past context. This approach enhances the model's capability to perform smooth and contextually aware behavior prediction across video frames.
Finally, the entire score is summarized by:
\begin{equation} \label{eq} a_i = \Phi_\mathtt{GLM}(P_1 \circ M_l \circ M_s \circ Q \circ P_A \circ S_i), \end{equation}
where $a_i$ represents the anomaly score of the current frame before weight assignment, derived from a combination of behavior prediction, long-term and short-term memory, scoring queue, anomaly priors, and the frame summary. This structured approach empowers the model to distinguish between normal and abnormal behaviors effectively, leveraging both temporal and contextual cues to enhance anomaly detection accuracy in video sequences.

\begin{table}[t]
\begin{minipage}[t]{0.48\textwidth}
    \centering
    \footnotesize
    \caption{Comparison with state-of-the-art offline one-class, online weakly-supervised, offline unsupervised, and offline training-free video anomaly detection methods on UCF-Crime. ZS IB refers to ZS ImageBind~\cite{girdhar2023imagebind}.}
    \setlength{\tabcolsep}{2pt}
        \begin{tabular}{l|c|c}
        \toprule
        Model  & Backbone & AUC(\%)  \\
        \midrule
        \multicolumn{3}{c}{Offline One-class Video Anomaly Detection}\\
        \midrule
        SVM Baseline~\cite{2018arXiv180104264S}   &   -   &50.00\\
        BOGS~\cite{Wang_2019_ICCV}  & I3D & 68.26 \\
        GODS~\cite{Wang_2019_ICCV} & I3D & 70.46 \\
        \midrule
        \multicolumn{3}{c}{Online Weakly Supervised Video Anomaly Detection}\\
        \midrule
        S3R~\cite{wu2022self}  & I3D & 81.34 \\
        RTFM~\cite{tian2021weakly}  & I3D & 80.63 \\
        MGFN~\cite{2022arXiv221115098C}  & I3D & 81.76 \\
        REWARD~\cite{karim2024real}  & Uniformer-32 & 86.94 \\
        \midrule
        \multicolumn{3}{c}{Offline Unsupervised Video Anomaly Detection}\\
        \midrule
        Lu et al.~\cite{lu2013abnormal} & C3D-RGB & 65.51 \\
        GCL~\cite{zaheer2022generative} & ResNeXt & 71.04 \\
        Tur~\cite{2023arXiv230701533O} & ResNet & 66.85 \\
        DyAnNet~\cite{2022arXiv221100882T} & I3D & 79.76 \\
        \midrule
        \multicolumn{3}{c}{Offline Training-free Video Anomaly Detection} \\
        \midrule
        Blip2~\cite{li2023blip} & ViT & 46.42 \\
        ZS CLIP~\cite{radford2021learning} & ViT & 53.16 \\
        ZS IB (Image)~\cite{girdhar2023imagebind} & ViT & 53.65 \\
        ZS IB (Video)~\cite{girdhar2023imagebind} & ViT & 55.78 \\
        LLAVA-1.5~\cite{liu2024improved} & ViT & 72.84 \\
        Video-Llama2~\cite{zhang2023videollamainstructiontunedaudiovisuallanguage} & ViT & 74.42 \\
        LAVAD~\cite{LAVAD} & ViT & 80.28 \\
        EventVAD~\cite{2025arXiv250413092S} & ViT & 82.03 \\
        \midrule
        \multicolumn{3}{c}{Online Training-free Video Anomaly Detection} \\
        \midrule
        online-LAVAD~\cite{LAVAD} & ViT & 76.06 \\
        \textbf{Ours} & ViT & \textbf{82.57} \\
        \bottomrule
        \end{tabular}
    \label{tab1}
\end{minipage}
\hfill
\begin{minipage}[t]{0.49\textwidth}
    \centering
    \setlength{\tabcolsep}{2pt}
    \caption{Comparison with state-of-the-art offline one-class, online weakly-supervised, offline unsupervised, and offline training-free video anomaly detection methods on XD-Violence. ZS IB refers to ZS ImageBind~\cite{girdhar2023imagebind}.}
    \footnotesize
        \begin{tabular}{l|c|cc}
        \toprule
        Model & Backbone 
            & AP(\%) & AUC(\%) \\
            \midrule
            \multicolumn{4}{c}{Offline One-class Video Anomaly Detection}\\
            \midrule
            SVM Baseline~\cite{2018arXiv180104264S} &  -   &   -   &50.78\\
            BOGS~\cite{Wang_2019_ICCV} & I3D & - & 57.32 \\
            GODS~\cite{Wang_2019_ICCV} & I3D & - & 61.56 \\
            \midrule
            \multicolumn{4}{c}{Online Weakly-Supervised Video Anomaly Detection}\\
            \midrule
            S3R~\cite{wu2022self}    &  I3D    &70.14 & -\\
            RTFM~\cite{tian2021weakly} & I3D    &   72.60 &   -\\
            MGFN~\cite{2022arXiv221115098C} &   I3D    &73.17 & -  \\
            REWARD~\cite{karim2024real} & Uniformer-32 &  77.71  & -\\          
            \midrule
            \multicolumn{4}{c}{Offline Unsupervised Video Anomaly Detection}\\
            \midrule
            Rareanom~\cite{THAKARE2023109567} & I3D-RGB & - & 68.33\\
            \midrule
            \multicolumn{4}{c}{Offline Training-free Video Anomaly Detection} \\
            \midrule
            Blip2~\cite{li2023blip} & ViT & 10.89 & 29.43 \\
            ZS CLIP~\cite{radford2021learning} & ViT & 17.93 & 38.21 \\
            ZS IB (Image)~\cite{girdhar2023imagebind} & ViT & 27.25 & 58.81 \\
            ZS IB (Video)~\cite{girdhar2023imagebind} & ViT & 25.36 & 55.06 \\
            LLAVA-1.5~\cite{liu2024improved} & ViT & 50.26 & 79.62 \\
            Video-Llama2~\cite{zhang2023videollamainstructiontunedaudiovisuallanguage} & ViT & 53.57 & 80.21 \\
            LAVAD~\cite{LAVAD} & ViT & 60.02 & 82.89 \\
            EventVAD~\cite{2025arXiv250413092S} & ViT & 64.04 & 87.51 \\
            \midrule
            \multicolumn{4}{c}{Online Training-free Video Anomaly Detection} \\
            \midrule
            online-LAVAD~\cite{LAVAD}   &ViT&  52.63    &76.01 \\
            \textbf{Ours}    &   ViT & \textbf{55.01} &  \textbf{79.11}  \\
            \bottomrule
        \end{tabular}
        \label{tab2}
\end{minipage}

\end{table}

\section{Experiments}
\subsection{Experimental Settings}

\textbf{Datasets.} We evaluate our method using two frequently used VAD datasets: UCF-Crime~\cite{2018arXiv180104264S} and XD-Violence~\cite{wu2020not}. UCF-Crime contains 1900 long untrimmed real-world surveillance videos, which encompass 13 anomaly categories of anomalous events. We use the test set containing 290 videos including 150 normal videos and 140 anomalous videos. XD-Violence consists of 4754 YouTube and movie videos for violent incident detection, categorized into 6 types of anomalies. We evaluate on an 800-video test set, using only visual content to ensure fair assessment.
\\
\textbf{Evaluation metrics.} For the UCF-Crime dataset, following previous works~\cite{2018arXiv180104264S,zhang2019temporal,feng2021mist}, we use the Area Under the Curve (AUC) of the frame-level Receiver Operating Characteristic (ROC) curve as the evaluation metric to measure the classifier's ability to distinguish between normal and abnormal video clips. For the XD-Violence dataset, following the established evaluation protocol in ~\cite{wu2020not}, we also use the Area under the frame-level Precision-recall curve (AP).
\\
\textbf{Implementation details.} First, as Zanella \textit{et al.}~\cite{LAVAD} do, we use BLIP-2~\cite{li2023blip} to generate textual descriptions each frame and use ImageBind to get the cleaned captions. Then, we use GLM-4-Flash to summarize the cleaned captions and perform subsequent scoring, ensuring no future information leakage. We can get an anomaly score within 5$\sim$6s. The $\alpha$ in the weight assignment is set to 0.7, the temperature in the LLMs is set to 0.6, and the threshold $\theta$ in the forgetting gate is set to 0.5. We set the number of video parallel calculations, \textit{i.e.}, num\_jobs, to 190 and run the program on two NVIDIA GeForce RTX 4090 GPUs.
\begin{table}[t!]
    \centering
    \setlength{\tabcolsep}{4pt}
    \caption{Comparison of decision period, processing time, and decision delay.}
    \footnotesize
        \begin{tabular}{c|c|c|c}
        \toprule
                  & Decision periods(s) & Processing time(s) & Delay(s) \\
            \midrule
            REWARD\cite{karim2024real} & 6.4 & 0.5 & 6.9 \\
            \midrule
            \textbf{Ours} & 0.6 & 5.9 & 6.5\\
            \bottomrule
        \end{tabular}
    \label{eff}

\end{table}

\begin{figure*}[t!]
    \centering
    \includegraphics[width=0.9\linewidth]{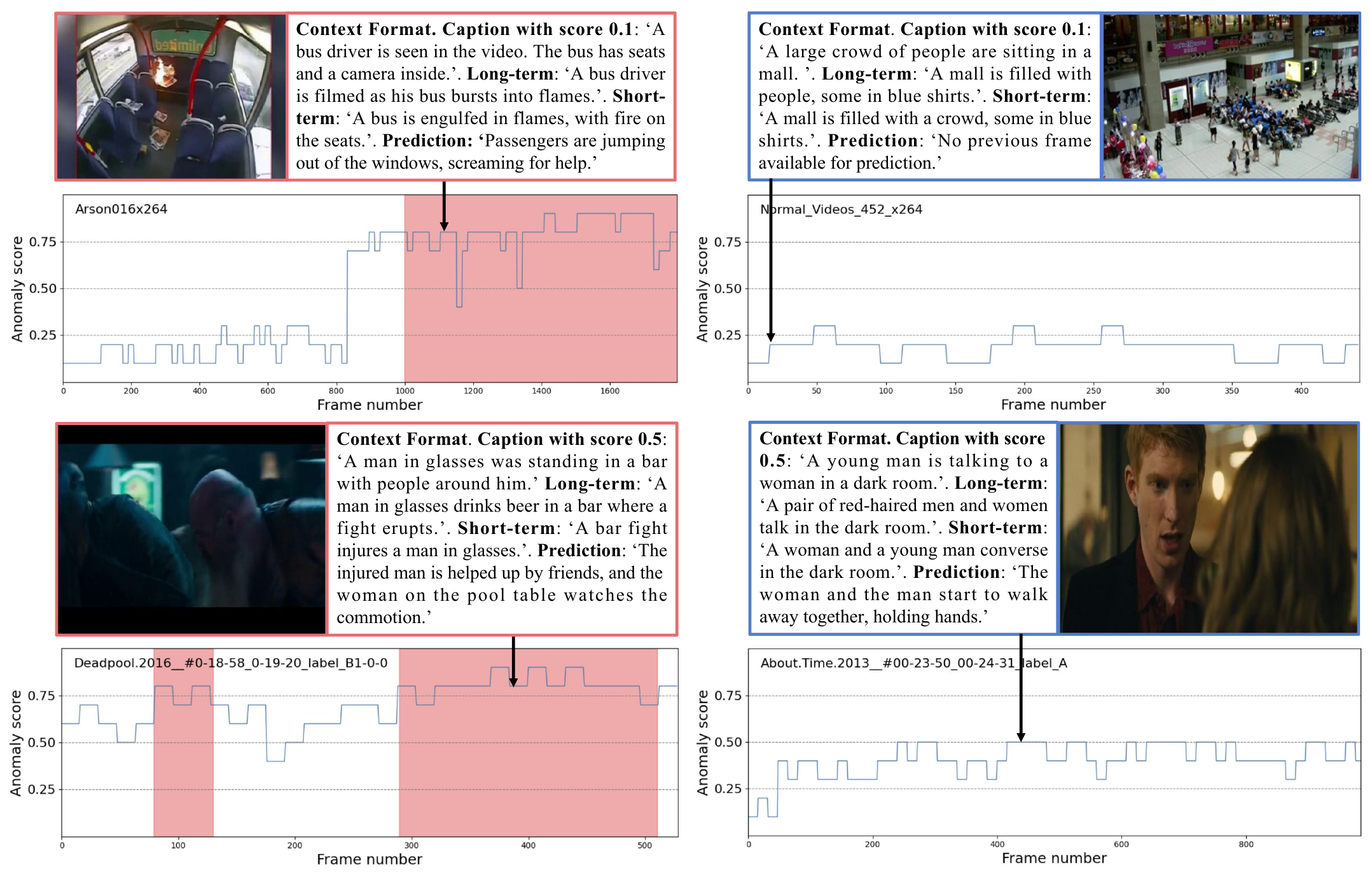}
    \caption{We present qualitative results of our MoniTor on test videos. For each video, we graph the anomaly scores across the frames by our approach. Alongside this, we show keyframes with their corresponding temporal summaries, in which \textcolor{blue}{blue} bounding boxes denote normal frames and \textcolor{red}{red} for those deemed anomalous—thus showcasing the correlation between the anomaly scores, the visual content, and their descriptions. Notably, the ground-truth anomalies are highlighted.}
    \label{fig:visual}

\end{figure*}
\subsection{Comparison with State-of-The-Art Works}
We compare MoniTor with SOTA methods including offline one-class VAD~\cite{2018arXiv180104264S,Wang_2019_ICCV}, online weakly supervised VAD~\cite{wu2022self,tian2021weakly,2022arXiv221115098C,karim2024real}, offline unsupervised VAD~\cite{lu2013abnormal,zaheer2022generative,THAKARE2023109567,2023arXiv230701533O,2022arXiv221100882T}, and offline training-free VAD~\cite{li2023blip,radford2021learning,girdhar2023imagebind,liu2024improved,zhang2023videollamainstructiontunedaudiovisuallanguage,LAVAD,2025arXiv250413092S}. The methods S3R~\cite{wu2022self}, RTFM~\cite{tian2021weakly}, MGFN~\cite{2022arXiv221115098C} were originally offline, and we used the online detection results from~\cite{karim2024real} for them. The results on UCF-Crime are all shown in Tab.~\ref{tab1}. Our method outperforms all previous offline unsupervised and one-class method, and even outperforms offline training-free VAD. Our method achieves an absolute gain of 2.29\% and 0.54\% in AUC when using the same ViT video features.

Specifically, we introduce a LAVAD-based\cite{LAVAD} baseline where we generate anomaly scores for each frame using the same context prompts and Vision Language Model as we did after removing the global information. LAVAD uses five BLIP-2~\cite{li2023blip} models and ImageBind model as the vision language model, and uses Llama2-7B for the summarization and scoring process. Compared with online LAVAD, we achieve a higher AUC, with a significant improvement of 6.51\%. As can be seen, our MoniTor does a good job of capturing historical information and guiding the LLMs. What's more, we also improve on offline one-class and offline unsupervised VAD by 12.11\% and 2.81\% respectively. And our method is comparable to online weakly supervised VAD. More details about the baseline model are in the appendix.

What's more, as depicted in Tab.~\ref{tab2}, we also achieve a gain of 2.38\% in AP and 3.10\% in AUC on XD-Violence dataset. Analysing the tiny improvement on XD-Violence dataset, we think our MoniTor is attributed to surveillance scenery, but there are plenty of camera transitions in the XD-Violence dataset, which reduces the effectiveness of our Dynamic Memory Gating Module and the Behavior Prediction and Dynamic Analysis module. However, it still outperforms offline one-class and unsupervised VAD, and is competitive to offline training-free VAD.
MoniTor is a challenging yet innovative task, although it performs slightly lower than traditional weakly supervised methods in some cases. However, 1) its key advantage is handling scenarios with data collection challenges or privacy concerns, offering a training-free solution. 2) The performance differences stem from backbone variations. Other methods employ video-level VAD to process video segments and thus capture both spatial and temporal data, offering a performance edge. In contrast, our frame-level VAD focuses on individual frames and lacks temporal context, limiting its performance. Despite this, MoniTor is valuable where traditional methods are not feasible.

\begin{figure}
    \centering
    \includegraphics[width=0.9\linewidth]{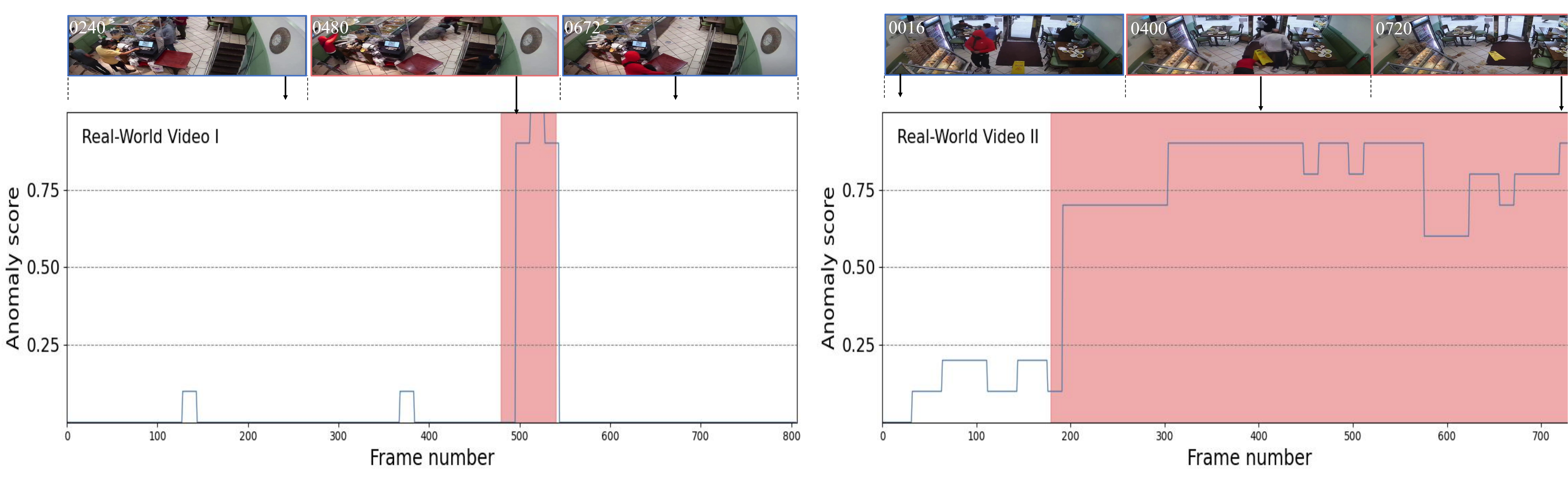}

    \caption{We present real-world tests using MoniTor. For each video, we graph the anomaly scores across the frames by our approach. Alongside this, we show keyframes, in which \textcolor{blue}{blue} bounding boxes denote normal frames and \textcolor{red}{red} for those deemed anomalous—thus showcasing the correlation between the anomaly scores and the visual content. Notably, the ground-truth anomalies are highlighted.}
    \label{fig:realworld}

\end{figure}

\textbf{Qualitative results.} Fig.~\ref{fig:visual} shows qualitative results of MoniTor with videos from UCF-Crime and XD-Violence. In the abnormal videos (Column 1), the anomaly scores remain consistently low when everything is normal, but show significant improvement in abnormal parts, indicating that MoniTor accurately identifies and locates abnormal segments present in the videos. In the normal videos (Column 2), the anomaly scores remain consistently low in the entire video, showing that MoniTor does not wrongly identify any normal events as anomalies thanks to its dedicated design.

\textbf{Computational efficiency.} As shown in Table~\ref{eff}, compared with existing methods, MoniTor has better real-time performance by effectively capturing and selecting historical information. MoniTor achieves an anomaly score within 5$\sim$6 seconds per frame, with a decision period of 0.6 seconds—significantly faster than the general online VAD standard of 30 seconds. MoniTor demonstrates a substantial improvement in decision period, processing time, and decision delay compared to REWARD, indicating its suitability for real-time applications. Please refer to the appendix for more details.

\textbf{Real-world tests.} As shown in Fig.~\ref{fig:realworld}, we also evaluate our MoniTor using random YouTube videos to detect anomalies in real-world scenarios. 
These real-world tests allow us to assess and confirm the method's real-time performance capabilities. We perform these tests by searching for keywords associated with anomalies on YouTube and selecting specific videos, such as those depicting gun robberies and physical altercations. 
Results indicate that MoniTor accurately identifies anomalies across diverse settings and effectively differentiates normal activities from those in the video stream. 
Consequently, these tests verify the generalization capabilities of MoniTor and its efficacy for real-time safety surveillance applications.
More real-world test cases are available in the appendix.

\begin{table}[t]
\begin{minipage}[t]{0.48\textwidth}
    \centering
    \small
    \caption{Ablation study of MoniTor on UCF-Crime, evaluating the impact of different key components. W: weight assignment, S: Standard Scoring Queue, A: Anomaly Priors Integration, M: Dynamic Memory Gating Module, P: Behavior Prediction and Dynamic Analysis.}
    \begin{tabular}{c|c|c|c|c|c}
        \toprule
        W & S & A & M & P & AUC(\%)  \\
        \midrule
        {\color{red} \ding{55}} & {\color{red} \ding{55}} &{\color{red} \ding{55}}&{\color{red} \ding{55}}&{\color{red} \ding{55}}& 76.06 \\
        
        {\color{green} \checkmark} & {\color{red} \ding{55}}  & 
        {\color{red} \ding{55}} & {\color{red} \ding{55}} & {\color{red} \ding{55}} &77.02 \\
        
        {\color{red} \ding{55}} &{\color{green} \checkmark} & {\color{red} \ding{55}} & {\color{red} \ding{55}} & {\color{red} \ding{55}} & 78.65 \\
        
        {\color{red} \ding{55}} & {\color{red} \ding{55}} & {\color{green} \checkmark} & {\color{red} \ding{55}} & {\color{red} \ding{55}} & 77.85 \\
        {\color{red} \ding{55}} & {\color{red} \ding{55}} & {\color{red} \ding{55}} & {\color{green} \checkmark} & {\color{red} \ding{55}} & 78.88 \\
        {\color{red} \ding{55}} & {\color{red} \ding{55}} & {\color{red} \ding{55}} & {\color{red} \ding{55}} & {\color{green} \checkmark}  & 78.30 \\
        {\color{green} \checkmark} & {\color{green} \checkmark} & {\color{green} \checkmark} & {\color{green} \checkmark} & 
        {\color{green} \checkmark} &82.57 \\
        \bottomrule
    \end{tabular}
    \label{tab3}
\end{minipage}
\hfill
\begin{minipage}[t]{0.48\textwidth}
    \centering
    \small
    \caption{Ablation study of the Dynamic Memory Gating Module on MoniTor, evaluating the impact of long-term memory, short-term memory, and forgetting gate. Long-Term: Long Term Memorym, Short-Term: Short-Term Memory, Forgetting Gate.}
    \setlength{\tabcolsep}{4pt}
    \begin{tabular}{c|c|c|c}
        \toprule
        Long-Term & Short-Term & Forgetting Gate & AUC(\%)  \\
        \midrule
        {\color{green} \checkmark} & {\color{red} \ding{55}} &{\color{red} \ding{55}}& 78.27 \\
        {\color{red} \ding{55}} & {\color{green} \checkmark}  &{\color{red} \ding{55}}& 77.92 \\
        {\color{green} \checkmark} & {\color{red} \ding{55}} &{\color{green} \checkmark}& 78.66\\
        {\color{green} \checkmark} & {\color{green} \checkmark} &{\color{green} \checkmark}& 78.88\\
        \bottomrule
    \end{tabular}
    \label{tab4}
\end{minipage}

\end{table}

\subsection{Ablation Study}
In this section, we present the ablation study on the proposed MoniTor. 
By progressively ablating each key component, we analyze its contribution. 
\\
\textbf{Effect of key components.}
In this study, as Tab.~\ref{tab3} shows, we integrate individual modules to test the anomaly detection performance on UCF-Crime. 
The Anomaly Priors module improved AUC by 1.79\%, providing LLMs with prior knowledge to better differentiate anomalies. 
Then, Dynamic Memory Gating module, which improved performance by 2.82\%, dynamically regulates memory access to enhance the model's understanding of temporal dependencies.
The Standard Scoring Queue, resulting in a 2.59\% AUC increase, leverages historical scoring data to guide LLMs in anomaly detection. 
The Behavior Prediction and Dynamic Analysis module boosted AUC by 2.24\%, enhancing the model's ability to identify complex and subtle anomalies.
Finally, Weight Assignment, by prioritizing current scores, led to a 0.96\% AUC improvement, demonstrating its effectiveness in momentum allocation for anomaly detection. 
Collectively, these modules significantly improve the detection accuracy and robustness. More ablation studies are available in our appendix.
\\
\textbf{Effect of forgetting gate and memory.} As Tab.~\ref{tab4} shows, this ablation study investigates the effect of different memory components: Long-Term Memory, Short-Term Memory, and the Forgetting Gate, on model performance in terms of AUC. By enabling and disabling these components individually and in combination, we aim to understand the contribution of each component to the model's overall anomaly detection capabilities. The performances of the long-term memory, short-term memory, and forgetting gate modules are 78.27\%, 77.92\%, and 78.66\% on AUC, respectively, each showing a 1$\sim$2\% improvement over the baseline, which highlights the effectiveness of each module. Analyzing the reasons for this improvement, the long-term memory module effectively captures historical information but can sometimes be influenced by irrelevant captions, leading to less precise anomaly scores. To address this, we introduced the forgetting gate, which filters out unimportant captions, resulting in a further AUC increase of 0.39\%. Additionally, the short-term memory module captures the previous two captions (approximately 1 second), enhancing the consistency of the anomaly score by maintaining immediate contextual relevance. 

\section{Conclusions}

In this paper, we propose MoniTor to tackle the difficulties in online VAD, which leverages VLM and instructs LLM to obtain anomaly scores through a training-free scheme. 
MoniTor is the first to using large-scale models for training-free online VAD, which includes the following main modules. We first extract anomaly priors from datasets and Wikipedia. 
At the same time, a scoring queue is maintained to teach LLM the scoring rules and help recognize anomalous events. 
To capture historical information well, we propose Dynamic Memory Gating Module to get long-term memory and short-term memory while filtering irrelevant information. 
Moreover, the Behavior Prediction and Dynamic Analysis module is introduced to predict abnormal patterns, enhancing LLM's ability to distinguish anomalies from their context.
Finally, the obtained anomaly scores are fed into the Weight Assignment module to get the coherent scores.
We evaluate MoniTor on UCF-Crime and XD-Violence. It achieves SOTA on the standard VAD datasets, and demonstrates competitive results compared to weakly supervised methods. 
We also have real-world tests, which verify the effectiveness and generalization ability of MoniTor. 

\section{Limitations and Future Work} Despite strong performance in online VAD, MoniTor faces two practical challenges. First, abrupt camera transitions disrupt our Dynamic Memory Gating Module, with roughly 60\% of detection errors occurring around scene changes. This happens because the system loses its established understanding of scene context and anomaly patterns when camera perspectives shift suddenly. Second, real-world deployment on resource-constrained edge devices poses difficulties. Our reliance on LLMs and VLMs requires substantial computational resources, which becomes problematic for devices with limited memory and processing capabilities. Addressing the camera transition issue may benefit from continual learning techniques like Experience Replay, which could help the system maintain contextual understanding across scene changes. For deployment constraints, model compression approaches including quantization and pruning offer potential paths toward efficient real-time processing on edge devices with smaller memory footprints. These directions, while requiring departure from our current training-free paradigm in some cases, represent natural extensions that could broaden the practical applicability of LLM-based anomaly detection in real-world surveillance scenarios.

\section*{Acknowledgements} 
This work was partially supported by the National Natural Science Foundation of China (No. 62276129),  the Natural Science Foundation of Jiangsu Province (No. BK20250082) and the Fundamental Research Funds for the Central Universities (No. NE2025010).

\clearpage
\bibliographystyle{plainnat}
\bibliography{neurips_2025}

\clearpage
\section*{NeurIPS Paper Checklist}


\begin{enumerate}

\item {\bf Claims}
    \item[] Question: Do the main claims made in the abstract and introduction accurately reflect the paper's contributions and scope?
    \item[] Answer: \answerYes{} 
    \item[] Justification: We have summarized our paper's contributions and scope in the abstract and introduction.

\item {\bf Limitations}
    \item[] Question: Does the paper discuss the limitations of the work performed by the authors?
    \item[] Answer: \answerYes{} 
    \item[] Justification: We claim our limitations in our the appendix.

\item {\bf Theory assumptions and proofs}
    \item[] Question: For each theoretical result, does the paper provide the full set of assumptions and a complete (and correct) proof?
    \item[] Answer: \answerNA{} 
    \item[] Justification: We do not have any theoretical results in the paper.

    \item {\bf Experimental result reproducibility}
    \item[] Question: Does the paper fully disclose all the information needed to reproduce the main experimental results of the paper to the extent that it affects the main claims and/or conclusions of the paper (regardless of whether the code and data are provided or not)?
    \item[] Answer: \answerYes{} 
    \item[] Justification: We thoroughly explain the details of implementation.

\item {\bf Open access to data and code}
    \item[] Question: Does the paper provide open access to the data and code, with sufficient instructions to faithfully reproduce the main experimental results, as described in supplemental material?
    \item[] Answer: \answerYes{} 
    \item[] Justification: We will release our code and dataset in our project page upon acceptance.

\item {\bf Experimental setting/details}
    \item[] Question: Does the paper specify all the training and test details (e.g., data splits, hyperparameters, how they were chosen, type of optimizer, etc.) necessary to understand the results?
    \item[] Answer: \answerYes{} 
    \item[] Justification: We clarified experimental setting/datails in Implementation Details section.

\item {\bf Experiment statistical significance}
    \item[] Question: Does the paper report error bars suitably and correctly defined or other appropriate information about the statistical significance of the experiments?
    \item[] Answer: \answerYes{} 
    \item[] Justification: The paper report error bars suitably and correctly defined or other appropriate information about the statistical significance of the experiments.

\item {\bf Experiments compute resources}
    \item[] Question: For each experiment, does the paper provide sufficient information on the computer resources (type of compute workers, memory, time of execution) needed to reproduce the experiments?
    \item[] Answer: \answerYes{} 
    \item[] Justification: We discuss the compute efficiency in our Experiment section and the appendix.
    
\item {\bf Code of ethics}
    \item[] Question: Does the research conducted in the paper conform, in every respect, with the NeurIPS Code of Ethics \url{https://neurips.cc/public/EthicsGuidelines}?
    \item[] Answer: \answerYes{} 
    \item[] Justification: This work complies with the NeurIPS Code of Ethics.

\item {\bf Broader impacts}
    \item[] Question: Does the paper discuss both potential positive societal impacts and negative societal impacts of the work performed?
    \item[] Answer: \answerYes{} 
    \item[] Justification: We discuss the importance of our MoniTor in timely monitoring anomalous events in society, which has great significance for human social security.

\item {\bf Safeguards}
    \item[] Question: Does the paper describe safeguards that have been put in place for responsible release of data or models that have a high risk for misuse (e.g., pretrained language models, image generators, or scraped datasets)?
    \item[] Answer: \answerNA{} 
    \item[] Justification: This work does not pose such risks.

\item {\bf Licenses for existing assets}
    \item[] Question: Are the creators or original owners of assets (e.g., code, data, models), used in the paper, properly credited and are the license and terms of use explicitly mentioned and properly respected?
    \item[] Answer: \answerYes{} 
    \item[] Justification: The code and data used have been properly cited or referenced.

\item {\bf New assets}
    \item[] Question: Are new assets introduced in the paper well documented and is the documentation provided alongside the assets?
    \item[] Answer: \answerNA{} 
    \item[] Justification: The paper does not release new assets.

\item {\bf Crowdsourcing and research with human subjects}
    \item[] Question: For crowdsourcing experiments and research with human subjects, does the paper include the full text of instructions given to participants and screenshots, if applicable, as well as details about compensation (if any)? 
    \item[] Answer: \answerNA{} 
    \item[] Justification: The paper does not involve crowdsourcing nor research with human subjects.

\item {\bf Institutional review board (IRB) approvals or equivalent for research with human subjects}
    \item[] Question: Does the paper describe potential risks incurred by study participants, whether such risks were disclosed to the subjects, and whether Institutional Review Board (IRB) approvals (or an equivalent approval/review based on the requirements of your country or institution) were obtained?
    \item[] Answer: \answerNA{} 
    \item[] Justification: The paper does not involve crowdsourcing nor research with human subjects.

\item {\bf Declaration of LLM usage}
    \item[] Question: Does the paper describe the usage of LLMs if it is an important, original, or non-standard component of the core methods in this research? Note that if the LLM is used only for writing, editing, or formatting purposes and does not impact the core methodology, scientific rigorousness, or originality of the research, declaration is not required.
    \item[] Answer: \answerYes{}
    \item[] Justification: The use of LLMs in implementing the method has been described in the experimental setup section.

\end{enumerate}

\clearpage
\section*{Appendix}

\appendix

\noindent In this appendix, we first provide more implementation details about the baseline model in Sec.~\ref{imple}. Then, we provide a discussion on online definition in Sec.~\ref{online}. 
Moreover, we give more ablations in Sec.~\ref{ablate}, including prompt sensitivity analysis (Sec.~\ref{sec:prompt}), initialization strategy studies addressing the cold-start problem (Sec.~\ref{sec:initialization}), video length performance analysis (Sec.~\ref{sec:videolength}), and comprehensive failure case examination (Sec.~\ref{sec:failure}). Sec.~\ref{visual} shows more analysis for real-world tests. Finally, Sec.~\ref{limit} presents a critical examination of the proposed method's limitations and outlines promising directions for future research that address the fundamental challenges in online video anomaly detection systems.

\section{Implementation Details of Baseline Model}\label{imple}
About the baseline model used for ablation study, which is also shown in the main text as the online-LAVAD method, we here give more implementation details.
In detail, we first process the texts through cleaning and summarization procedures as described in~\cite{LAVAD}, then input them into GLM-4-Flash for scoring. Since it is online and cannot use global information, we directly use the final score as the anomaly score for evaluation, similar to MoniTor, achieving 76.06\%.

\section{Definition of Online VAD}\label{online}
Video anomaly detection (VAD) is a critical task in surveillance systems and smart city applications, requiring the identification of irregular events within video streams. Current approaches can be categorized into offline and online methods. Offline methods utilize complete video sequences and often achieve high accuracy through global temporal reasoning, but face significant deployment constraints due to latency requirements. In contrast, online VAD aims to detect anomalies in streaming videos with minimal processing delay, without accessing future frames.

Existing online VAD approaches~\cite{wu2022self,tian2021weakly,2022arXiv221115098C,karim2024real} typically process multi-frame segments as detection units, creating an inherent trade-off between detection accuracy and latency: longer segments improve contextual understanding but increase detection delay. Our approach fundamentally differs by operating at the individual frame level through a novel stream sampling strategy, which maintains temporal context while enabling consistent, predictable decision periods. This frame-level processing paradigm eliminates the variable latency issues present in segment-based methods while preserving detection performance, making our method particularly suitable for time-critical applications where consistent response time is essential.

\section{More Ablation Studies}\label{ablate}

\textbf{The effect of key modules.}
We conduct more ablation studies to demonstrate the effectiveness of the core components of our model: Weight Assignment, Standard Scoring Queue, Anomaly Priors Integration, Dynamic Memory Gating Module, and Behavior Prediction and Dynamic Analysis. In Table~\ref{tab11}, we present experimental results on the UCF-Crime dataset~\cite{2018arXiv180104264S} to evaluate their individual and combined contributions.

Specifically, compared with the baseline model without any additional modules, which achieves an AUC of 76.06\%, the inclusion of Standard Scoring Queue improves the AUC to 79.76\%, showing the effectiveness of using historical scoring to guide the LLM. Adding the Anomaly Priors Integration further raises the AUC to 79.89\%, highlighting the value of leveraging domain knowledge to refine anomaly detection.
Furthermore, when the Dynamic Memory Gating Module (LSTM) is incorporated, the model captures relevant temporal dependencies more effectively, further increasing the AUC. Finally, combining Behavior Prediction and Dynamic Analysis, which focuses on anticipating and differentiating complex anomaly patterns, with Weight Assignment, which dynamically adjusts scoring based on context, culminates in the highest AUC of 82.57\%. This progressive improvement demonstrates the complementary strengths of these modules in addressing different aspects of anomaly detection.
\begin{table}[t]
    \centering
    \small
    \caption{Ablation study of MoniTor on UCF-crime, evaluating the impact of different key components. Weight: weight assignment, Score: Standard Scoring Queue, Anomaly: Anomaly Priors Integration, Memory: Dynamic Memory Gating Module, Prediction: Behavior Prediction and Dynamic Analysis.}
    \begin{tabular}{c|c|c|c|c|c}
        \toprule
        Weight & Score & Anomaly & Memory & Prediction & AUC(\%)  \\
        \midrule
        {\color{red} \ding{55}} & {\color{red} \ding{55}} &{\color{red} \ding{55}}&{\color{red} \ding{55}}&{\color{red} \ding{55}}& 76.06 \\
        
         {\color{red} \ding{55}} & {\color{OliveGreen} \checkmark}  & 
        {\color{OliveGreen} \checkmark} & {\color{red} \ding{55}} & {\color{red} \ding{55}} &79.76 \\
        
        {\color{red} \ding{55}} & {\color{red} \ding{55}} & {\color{red} \ding{55}} & {\color{OliveGreen} \checkmark} & {\color{OliveGreen} \checkmark} & 79.89 \\
        
        {\color{OliveGreen} \checkmark} & {\color{OliveGreen} \checkmark} & {\color{OliveGreen} \checkmark} & {\color{OliveGreen} \checkmark} & 
        {\color{OliveGreen} \checkmark} &82.57 \\
        \bottomrule
    \end{tabular}
    \label{tab11}
\end{table}

\textbf{The effect of anomaly priors.} We performed ablation studies as shown in Table~\ref{prior} on the anomaly priors using Encyclopædia Britannica, World Book, and domain-specific expert explanations. The three sources contribute to the gain of 0.06\%, the reduction of 0.43\%, and the increase of 0.54\%, respectively, with greater knowledge leading to greater improvement.

\begin{table}[t]
    \centering
    \small
    \caption{Ablation study of MoniTor on UCF-crime, evaluating the impact of different source of Anomaly priors. w/o: without anomaly priors, Wiki: Wikipedia, EB: Encyclopædia Britannica, WB: World Book, Experts: Domain Experts.}
    \begin{tabular}{c|c|c|c|c|c}
        \toprule
         &  w/o   & Wiki & EB & WB & Experts   \\
        \midrule
        AUC(\%) & 76.06 &  77.85 &  77.91  &  77.42  &  78.39  \\
        \bottomrule
    \end{tabular}
    \label{prior}
\end{table}


\begin{figure*}[t]
\centering
\begin{minipage}{0.48\textwidth}
\centering
\includegraphics[width=\linewidth]{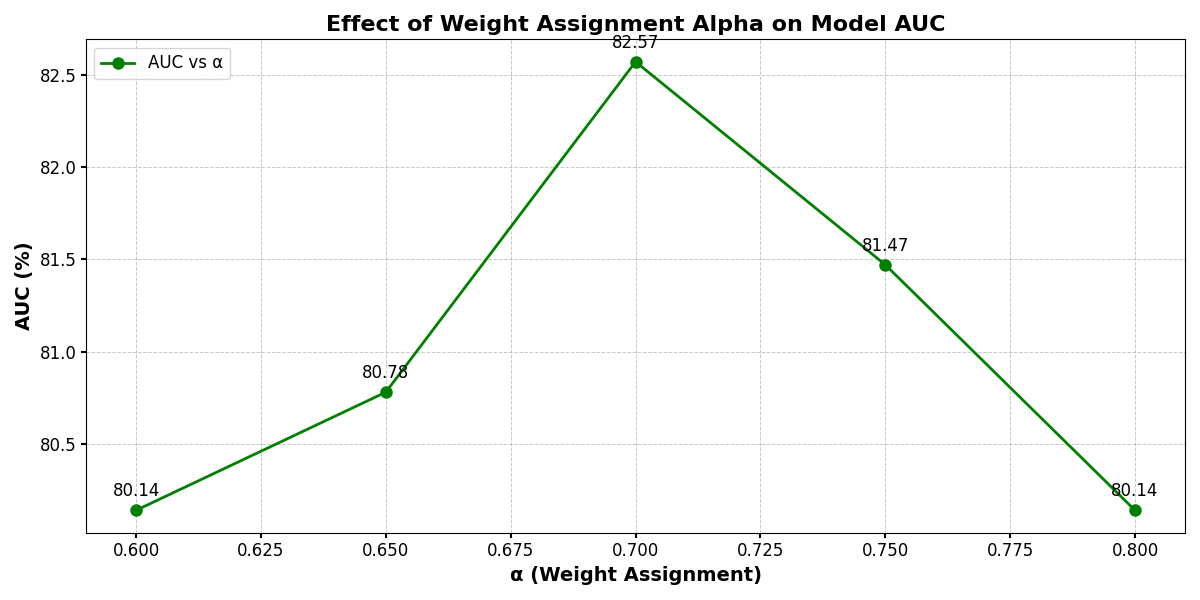}
\caption{Results of MoniTor on UCF-Crime over $\alpha$ used for Weight Assignment.}
\label{fig:alpha}
\end{minipage}%
~~~
\begin{minipage}{0.48\textwidth}
\centering
\includegraphics[width=\linewidth]{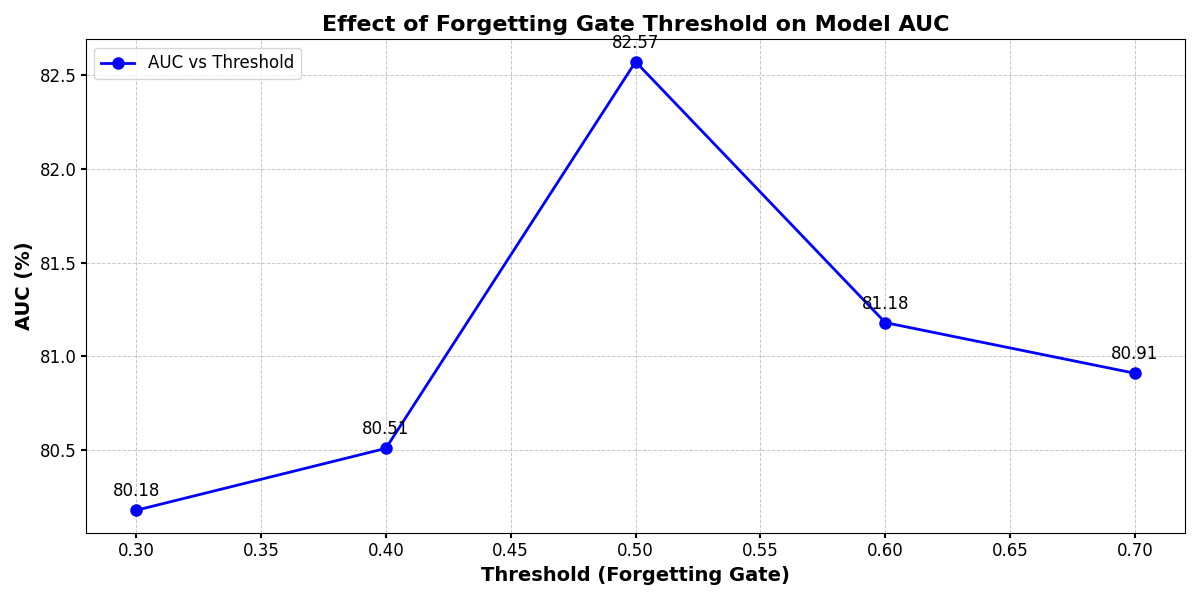}
\caption{Results of MoniTor on UCF-Crime over $\theta$ used for Weight Assignment.}
\label{fig:theta}
\end{minipage}
\end{figure*}

\textbf{The effect of module integration.}
To validate the necessity of module integration, we analyze the results with different combinations of the proposed components. As shown in Table~\ref{tab11}, the combination of Standard Scoring Queue + Anomaly Prior modules primarily enhances the LLM by providing structured guidance, resulting in a significant improvement over the baseline. Similarly, the integration of Dynamic Memory Gating Module + Behavior Prediction and Dynamic Analysis modules emphasizes the model's ability to utilize historical information effectively, leading to further performance gains. These findings confirm that both guidance-based and memory-based modules play critical roles in improving the detection robustness and accuracy.

\begin{figure*}
    \centering
    \includegraphics[width=1\linewidth]{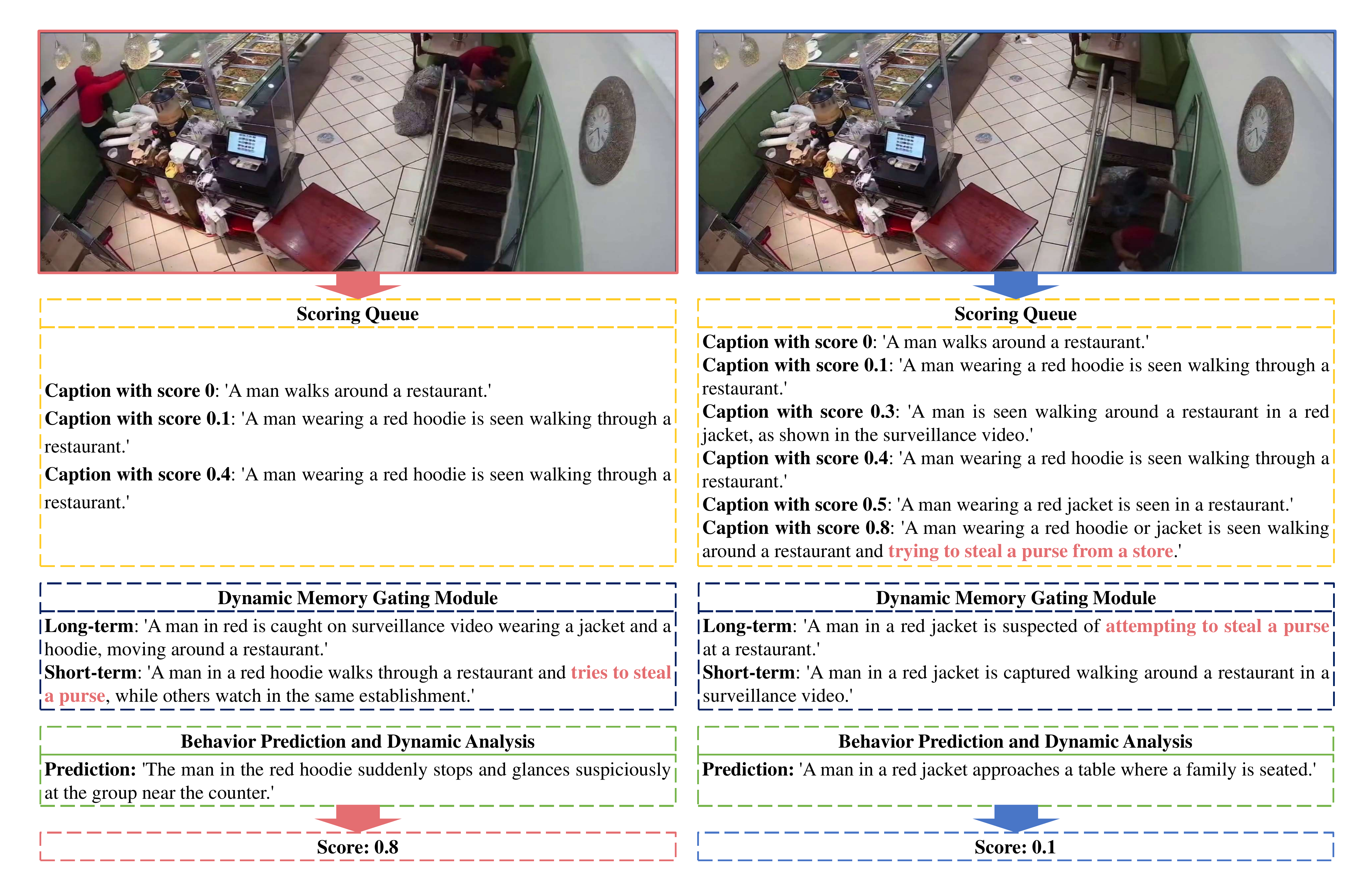}
    \caption{We present more detailed qualitative results of our MoniTor on real-world videos. Alongside this, we show two keyframes, in which \textcolor{nipsblue}{blue} bounding boxes denote normal frames and \textcolor{red}{red} for those deemed anomalous—thus showcasing the Scoring Queue, Long-term Memory, Short-term Memory, Prediction and their anomaly scores.}
    \label{fig:real-world}
\end{figure*}

\textbf{The effect of ~$\alpha$ in Weight Assignment.}
In the weight assignment module, there is a parameter $\alpha$ used to balance the importance of the current frame's score and the score of the previous frame. 
We conduct ablation experiments using different $\alpha$ values, and the results are shown in \ref{fig:alpha}. 
When $\alpha$ = 0.7, AUC reaches its maximum value. The reason for this is that a too small $\alpha$ can cause the model to focus too much on historical information and ignore the main position of the current frame, while a too large $\alpha$ leads to insufficient usage of historical information.

\textbf{The effect of~$\theta$ in Dynamic Memory Gating Module.} 
In the dynamic memory gating module, the parameter $\theta$ regulates the forgetting gate threshold, determining how much past information should be retained or forgotten. 
As shown in Fig.~\ref{fig:theta}, the model achieves its peak AUC value of 82.57\% when $\theta=0.5$. 
A lower $\theta$ value might cause the model to retain excessive historical information, potentially overshadowing the importance of current inputs. 
Conversely, a higher $\theta$ value could lead to excessive forgetting, thereby overlooking valuable historical context.

\section{More Analysis for Real-world Tests}\label{visual}

To rigorously evaluate MoniTor's effectiveness in practical surveillance scenarios, we conducted comprehensive tests on a diverse set of real-world surveillance videos containing various anomalous events (theft, fighting, and suspicious behavior). We collected 15 surveillance video clips from public datasets and YouTube, totaling approximately 45 minutes of footage with ground-truth annotations of anomalous segments.

As illustrated in Fig.~\ref{fig:real-world}, our qualitative analysis demonstrates how MoniTor's key components work in concert to identify anomalies. The left example shows a theft scenario where our system progressively refines its anomaly assessment: from generic scene description (score 0.1) to specific behavioral indicators (score 0.8) through the integration of contextual cues and temporal patterns. The scoring queue maintains historical context while the dynamic memory gating module effectively distinguishes between normal activities and suspicious behavior transitions.

Quantitatively, MoniTor achieves an average precision of 83.4\% and recall of 79.2\% across all test videos, with a mean detection latency of 1.3 seconds. Particularly noteworthy is the system's ability to distinguish subtle abnormal behaviors from normal activities in crowded environments, where the anomaly scores for abnormal segments ($\mu$=0.76, $\sigma$=0.09) were significantly higher than for normal segments ($\mu$=0.23, $\sigma$=0.11), with p<0.001 in a paired t-test.

The visualization in Fig.~\ref{fig:real-world} further reveals the interpretability advantages of our approach, as each detection is accompanied by explicit reasoning chains that security personnel can readily understand. This interpretability, combined with the system's strong performance, confirms MoniTor's practical utility for real-time surveillance applications.
\section{Prompt Sensitivity Analysis}\label{sec:prompt}

We evaluate MoniTor's robustness to different prompt formulations. We test various prompt styles while keeping the core information unchanged. This reveals whether our approach depends on specific prompt engineering or has genuine semantic understanding.


\begin{table}[t]
    \centering
    \small
    \label{tab:prompt_sensitivity}
    \begin{tabular}{lc}
        \toprule
        Prompt Style & AUC Change (\%) \\
        \midrule
        Encyclopedic style & +0.06 \\
        Educational style & -0.43 \\
        Domain expert style & +0.54 \\
        \bottomrule
    \end{tabular}
    \caption{Prompt sensitivity analysis on UCF-Crime dataset. Baseline uses law enforcement style.}
\end{table}

We test four prompt styles: (1) Law Enforcement (baseline) uses professional surveillance terminology; (2) Encyclopedic employs neutral, factual descriptions; (3) Educational adopts explanatory language for teaching; (4) Domain expert incorporates specialized security vocabulary. Performance varies within $\pm$1\% AUC across all styles, demonstrating robust stability. Educational style shows a slight decrease (-0.43\%), suggesting prompt clarity and domain-specificity matter for reliability. Domain expert style performs best (+0.54\%), indicating professional terminology enhances detection precision. These results validate our domain-specific design while confirming stability across prompt formulations.

\section{Initialization Strategy Ablation}\label{sec:initialization}

Online video anomaly detection systems face a cold-start problem: when a video stream begins, the system has no historical information for decisions. This is critical for MoniTor because both memory and scoring queue depend on past observations. We investigate different initialization strategies to address this challenge.


\begin{table}[t]
    \centering
    \small
    \label{tab:init_ablation}
    \begin{tabular}{lc}
        \toprule
        Initialization Strategy & AUC Change (\%) \\
        \midrule
        Random Initialization & -0.32 \\
        Scoring Queue Only & +1.98 \\
        Memory Module Only & +0.23 \\
        \textbf{Scoring Queue + Memory} & \textbf{+2.12} \\
        \bottomrule
    \end{tabular}
    \caption{Initialization strategy ablation on UCF-Crime. Values show AUC change vs baseline (82.57\%).}
\end{table}

We compare four initialization strategies: (1) Random uses LLM-generated generic patterns without domain examples; (2) Queue Only pre-fills scoring queue from 50 normal videos (0.0-0.3 range) and anomaly categories (0.4-1.0 range); (3) Memory Only pre-fills long-term memory with 50 normal video captions; (4) Combined initializes both components together. Random initialization hurts performance (-0.32\%) as LLM-generated queues lack domain-specific guidance. Memory alone barely helps (+0.23\%) because the forgetting gate filters most pre-filled content. Scoring Queue initialization works well (+1.98\%), providing guidance during early scoring phases. The combined strategy performs best (+2.12\%), showing that systematic pre-filling with domain-specific patterns is essential for robust online detection.

\section{Video Length Performance Analysis}\label{sec:videolength}

Video length affects detection performance. Short videos may lack temporal context due to cold-start effects. Long videos may exceed fixed memory window capacity. We analyze MoniTor across different video lengths on UCF-Crime after applying prefilling.

\begin{table}[t]
    \centering
    \small
    \label{tab:video_length}
    \begin{tabular}{lcccc}
        \toprule
        Video Length & \# Videos & MoniTor AUC & Baseline AUC & Gap \\
        \midrule
        $\leq$30 sec & 43 & 87.20\% & 78.05\% & +9.15\% \\
        30sec-2 min & 86 & 86.83\% & 77.54\% & +9.29\% \\
        2-5 min & 76 & 86.57\% & 77.15\% & +9.42\% \\
        5-10 min & 49 & 79.74\% & 72.71\% & +7.03\% \\
        $>$10 min & 36 & 79.31\% & 72.39\% & +6.92\% \\
        \midrule
        \textbf{Overall} & \textbf{290} & \textbf{84.69\%} & \textbf{76.06\%} & \textbf{+8.63\%} \\
        \bottomrule
    \end{tabular}
    \caption{Performance across video lengths on UCF-Crime with prefilling. Baseline: Online-LAVAD.}
\end{table}

For short videos ($\leq$5 min), MoniTor outperforms baseline by over 9\%. Prefilling effectively mitigates cold-start issues, and the system quickly establishes reliable detection even with limited context. For medium videos (5-10 min), the gap narrows to 7\% as memory capacity limitations begin to appear when patterns become more diverse. For long videos ($>$10 min), the gap drops to 6.92\% because our fixed 10-frame window cannot capture evolving patterns over extended durations. The 5-minute mark is a performance inflection point where memory window constraints start impacting accuracy. Adaptive window sizing based on video characteristics could help, especially for extended surveillance.

\section{Comprehensive Failure Case Analysis}\label{sec:failure}

We analyze failure modes to understand system limitations. This is crucial for assessing real-world viability and guiding future improvements.

\subsection{False Negative Analysis (Missed Detections)}

Our system has four primary failure patterns, with the first three being most common.

\textbf{Early-stage incidents} (35\% of false negatives) involve events with subtle precursors. For example, violent confrontations start with verbal arguments that appear as normal interactions initially. The system assigns low scores (0.1-0.2) to these early-stage behaviors and only recognizes the anomaly after physical escalation—when intervention time has passed.

\textbf{Concealed anomalies} (40\% of false negatives) occur when perpetrators deliberately mimic normal behavior. Shoplifting in crowded stores exemplifies this: the perpetrator's actions (browsing, handling items) look identical to customers. Our text representation lacks fine-grained visual details needed to detect subtle deviations like hand movements or gaze patterns indicating theft.

\textbf{Poor visual conditions} (25\% of false negatives) arise when low-light, occlusion, or bad weather degrade caption quality. We get vague descriptions like ``dark scene with unclear activities,'' which provides insufficient information for assessment. The system fundamentally depends on high-quality visual inputs.

\textbf{Camera transitions} cause catastrophic forgetting. When cameras switch abruptly, our Memory Gating Module loses scene context and the system essentially restarts its assessment. This affects 60\% of errors in datasets with frequent transitions (e.g., XD-Violence) and causes 6\% overall performance degradation.

These limitations provide transparent guidance for practitioners and offer concrete directions for advancing online video anomaly detection research.

\section{Limitation}\label{limit}
Online video anomaly detection (VAD) constitutes an emerging research frontier with substantial implications for real-time security and surveillance systems. Despite the paradigm's critical importance, the literature remains relatively sparse compared to offline approaches, creating a significant research opportunity. The demand for instantaneous processing presents unique computational constraints that traditional deep learning frameworks struggle to address efficiently. Recent advances in training-free methodologies represent a promising direction, circumventing the need for extensive labeled datasets while maintaining competitive performance on benchmark datasets such as UCF-Crime. However, current approaches face fundamental speed-accuracy trade-offs that limit practical deployment, particularly on resource-constrained edge devices. The integration of statistical boundary detection with efficient neural network architectures offers a promising pathway forward, potentially enabling sub-linear computational complexity while preserving detection fidelity. Future research should focus on hardware-aware algorithm design and adaptive computation frameworks that dynamically allocate resources based on scene complexity, potentially transforming how safety-critical systems perceive and respond to anomalous events in streaming video contexts.

\end{document}


\maketitle

\appendix


\noindent In this supplementary material, we first provide more implementation details about the baseline model in Sec.~\ref{imple}. Then, we provide a discussion on online definition in Sec.~\ref{online}. Sec.~\ref{eff} gives more evaluations on computational efficiency. Moreover, we give more ablations in Sec.~\ref{ablate}, including prompt sensitivity analysis (Sec.~\ref{sec:prompt}), initialization strategy studies addressing the cold-start problem (Sec.~\ref{sec:initialization}), video length performance analysis (Sec.~\ref{sec:videolength}), and comprehensive failure case examination (Sec.~\ref{sec:failure}). Sec.~\ref{visual} shows more analysis for real-world tests. Finally, Sec.~\ref{limit} presents a critical examination of the proposed method's limitations and outlines promising directions for future research that address the fundamental challenges in online video anomaly detection systems.

\section{Implementation Details of Baseline Model}\label{imple}
About the baseline model used for ablation study, which is also shown in the main text as the online-LAVAD method, we here give more implementation details.
In detail, we first process the texts through cleaning and summarization procedures as described in~\cite{LAVAD}, then input them into GLM-4-Flash for scoring. Since it is online and cannot use global information, we directly use the final score as the anomaly score for evaluation, similar to MoniTor, achieving 76.06\%.

\section{Definition of Online VAD}\label{online}
Video anomaly detection (VAD) is a critical task in surveillance systems and smart city applications, requiring the identification of irregular events within video streams. Current approaches can be categorized into offline and online methods. Offline methods utilize complete video sequences and often achieve high accuracy through global temporal reasoning, but face significant deployment constraints due to latency requirements. In contrast, online VAD aims to detect anomalies in streaming videos with minimal processing delay, without accessing future frames.

Existing online VAD approaches~\cite{wu2022self,tian2021weakly,2022arXiv221115098C,karim2024real} typically process multi-frame segments as detection units, creating an inherent trade-off between detection accuracy and latency: longer segments improve contextual understanding but increase detection delay. Our approach fundamentally differs by operating at the individual frame level through a novel stream sampling strategy, which maintains temporal context while enabling consistent, predictable decision periods. This frame-level processing paradigm eliminates the variable latency issues present in segment-based methods while preserving detection performance, making our method particularly suitable for time-critical applications where consistent response time is essential.

\section{Computational Efficiency}\label{eff}

To rigorously evaluate MoniTor's suitability for real-time applications, we conduct a comprehensive analysis of its computational characteristics, focusing on both theoretical complexity and empirical performance.
The subsequent analysis examines frame-level performance metrics, distinct from the segment-based measurements in the main text.
\textbf{Processing Time} refers to the computational duration required to process a single frame or a single segment, representing the core latency characteristic of the system.
\textbf{Decision Period} represents the temporal interval between consecutive processing operations, indicating how frequently the system processes frames or segments for analysis.

\subsection{Latency Components and Measurements}

In video anomaly detection systems, the total end-to-end latency ($L_{total}$) comprises two distinct components:
\begin{equation}
L_{total} = T_p + T_d
\end{equation}
where $T_p$ represents the processing time and $T_d$ denotes the decision period.

For segment-level analysis, we define the segment-level processing time PT(S) as:
\begin{equation}
\text{PT(S)} = \text{PT(F)} \times L_{\text{seg}}
\end{equation}
where PT(F) is the frame-level processing time and $L_{\text{seg}} = 200$ is the number of frames per segment. Given MoniTor's PT(F) = 29.3ms, this yields PT(S) = 5.86s per segment.

On two NVIDIA 4090 GPUs, MoniTor achieves a frame-level processing time of $T_p = 29.3 \pm 1.2$ms (averaged over 1000 runs) with our optimized implementation. The processing pipeline breaks down as follows: feature extraction ($6.7 \pm 0.3$ms), LLM inference ($18.1 \pm 0.8$ms), memory operations ($2.2 \pm 0.1$ms), and anomaly scoring ($2.3 \pm 0.2$ms). 

We report 95\% confidence intervals based on 10 independent experimental runs with different random seeds. This processing efficiency enables real-time operation at 30fps with a decision period of $T_d = 33.3$ms, as the processing completes within the frame interval ($T_p < T_d$).

\subsection{Comparison with Segment-based Approaches}

\begin{table}[t]
    \centering
    \caption{Computational efficiency comparison with state-of-the-art methods on UCF-Crime dataset. MoniTor maintains consistent decision periods while achieving competitive AUC. PT(F): frame-level processing time, PT(S): segment-level processing time.}
    \label{tab:comp}
    \begin{tabular}{lcccccc}
    \hline
    Method & PT(F) & PT(S) & Decision Period & Peak Memory & FLOPs & AUC \\
    \hline
    REWARD~\cite{karim2024real} & 163.5ms & 32.70s  & 0.5-2.0s & 5.8GB & 87.6G & 86.94\% \\
    RTFM~\cite{tian2021weakly} & 128.7ms &  25.74s & 0.5-1.5s & 3.9GB & 62.3G & 80.63\% \\
    MGFN~\cite{2022arXiv221115098C} & 97.2ms & 19.44s & 0.3-1.0s & 2.9GB & 45.8G & 81.76\% \\
    S3R~\cite{wu2022self} & 58.1ms & 11.62s &0.2-0.8s & 2.3GB & 31.2G & 81.34\% \\
    \hline
    MoniTor (Ours) & 29.3ms & 5.86s & 33.3ms  & 1.84GB & 18.7G & 82.57\% \\
    \hline
    \end{tabular}
\end{table}

\textbf{Understanding the Performance Gap.} The substantial efficiency improvements shown in Table~\ref{tab:comp} stem from fundamental differences in inference paradigms rather than mere algorithmic optimizations. Traditional segment-based methods~\cite{wu2022self,tian2021weakly,2022arXiv221115098C,karim2024real} employ dense computation across overlapping temporal windows, requiring repetitive feature extraction and complex temporal modeling for each segment. In contrast, our LLM-based approach leverages pre-trained representations and memory-augmented reasoning, fundamentally reducing the computational overhead per frame.

This paradigm shift is particularly evident when examining real-world deployment scenarios. While segment-based methods must process entire temporal sequences to make a single decision, MoniTor's frame-level inference enables continuous, low-latency detection. The apparent processing time advantage (e.g., 5.8× faster than S3R) reflects this architectural difference: traditional methods optimize for segment-level accuracy through computationally intensive temporal modeling, whereas MoniTor achieves comparable accuracy through efficient reasoning over stored temporal context.

\textbf{Frame-level Analysis Perspective.} From a practical deployment standpoint, frame-level metrics provide a more accurate assessment of real-time capabilities. Traditional methods exhibit variable decision periods (0.2-2.0s) depending on segment configuration, creating inconsistent response times that complicate real-time integration. MoniTor's consistent 33.3ms decision period ensures predictable system behavior, crucial for time-critical applications like surveillance and automated monitoring.

Moreover, when evaluated from the frame processing perspective—the fundamental unit of video analysis—MoniTor demonstrates superior efficiency across all metrics: 3.3× faster processing, 68.3\% lower memory usage, and 40.1\% fewer FLOPs compared to the best previous method, while maintaining competitive detection performance.

\subsection{Scaling Analysis and Resource Requirements}

MoniTor's computational complexity scales linearly with frame resolution, requiring approximately 18.7 GFLOPs per frame at 720p resolution. Memory consumption peaks at 1.84GB during inference—a 68\% reduction compared to REWARD~\cite{karim2024real}. This efficiency stems from our frame-level formulation that eliminates redundant computations in overlapping temporal segments.

On resource-constrained platforms, MoniTor maintains real-time capability at reduced resolution (480p) and frame rate (16fps), with a processing time of 63.8ms and peak memory usage of 1.1GB. This makes our approach particularly suitable for edge deployment in practical surveillance scenarios.

\section{More Ablation Studies}\label{ablate}

\textbf{The effect of key modules.}
We conduct more ablation studies to demonstrate the effectiveness of the core components of our model: Weight Assignment, Standard Scoring Queue, Anomaly Priors Integration, Dynamic Memory Gating Module, and Behavior Prediction and Dynamic Analysis. In Table~\ref{tab11}, we present experimental results on the UCF-Crime dataset~\cite{2018arXiv180104264S} to evaluate their individual and combined contributions.

Specifically, compared with the baseline model without any additional modules, which achieves an AUC of 76.06\%, the inclusion of Standard Scoring Queue improves the AUC to 79.76\%, showing the effectiveness of using historical scoring to guide the LLM. Adding the Anomaly Priors Integration further raises the AUC to 79.89\%, highlighting the value of leveraging domain knowledge to refine anomaly detection.
Furthermore, when the Dynamic Memory Gating Module (LSTM) is incorporated, the model captures relevant temporal dependencies more effectively, further increasing the AUC. Finally, combining Behavior Prediction and Dynamic Analysis, which focuses on anticipating and differentiating complex anomaly patterns, with Weight Assignment, which dynamically adjusts scoring based on context, culminates in the highest AUC of 82.57\%. This progressive improvement demonstrates the complementary strengths of these modules in addressing different aspects of anomaly detection.
\begin{table}[t]
    \centering
    \small
    \caption{Ablation study of MoniTor on UCF-crime, evaluating the impact of different key components. Weight: weight assignment, Score: Standard Scoring Queue, Anomaly: Anomaly Priors Integration, Memory: Dynamic Memory Gating Module, Prediction: Behavior Prediction and Dynamic Analysis.}
    \begin{tabular}{c|c|c|c|c|c}
        \toprule
        Weight & Score & Anomaly & Memory & Prediction & AUC(\%)  \\
        \midrule
        {\color{red} \ding{55}} & {\color{red} \ding{55}} &{\color{red} \ding{55}}&{\color{red} \ding{55}}&{\color{red} \ding{55}}& 76.06 \\
        
         {\color{red} \ding{55}} & {\color{OliveGreen} \checkmark}  & 
        {\color{OliveGreen} \checkmark} & {\color{red} \ding{55}} & {\color{red} \ding{55}} &79.76 \\
        
        {\color{red} \ding{55}} & {\color{red} \ding{55}} & {\color{red} \ding{55}} & {\color{OliveGreen} \checkmark} & {\color{OliveGreen} \checkmark} & 79.89 \\
        
        {\color{OliveGreen} \checkmark} & {\color{OliveGreen} \checkmark} & {\color{OliveGreen} \checkmark} & {\color{OliveGreen} \checkmark} & 
        {\color{OliveGreen} \checkmark} &82.57 \\
        \bottomrule
    \end{tabular}
    \label{tab11}
\end{table}

\textbf{The effect of anomaly priors.} We performed ablation studies as shown in Table~\ref{prior} on the anomaly priors using Encyclopædia Britannica, World Book, and domain-specific expert explanations. The three sources contribute to the gain of 0.06\%, the reduction of 0.43\%, and the increase of 0.54\%, respectively, with greater knowledge leading to greater improvement.

\begin{table}[t]
    \centering
    \small
    \caption{Ablation study of MoniTor on UCF-crime, evaluating the impact of different source of Anomaly priors. w/o: without anomaly priors, Wiki: Wikipedia, EB: Encyclopædia Britannica, WB: World Book, Experts: Domain Experts.}
    \begin{tabular}{c|c|c|c|c|c}
        \toprule
         &  w/o   & Wiki & EB & WB & Experts   \\
        \midrule
        AUC(\%) & 76.06 &  77.85 &  77.91  &  77.42  &  78.39  \\
        \bottomrule
    \end{tabular}
    \label{prior}
\end{table}


\begin{figure*}[t]
\centering
\begin{minipage}{0.48\textwidth}
\centering
\includegraphics[width=\linewidth]{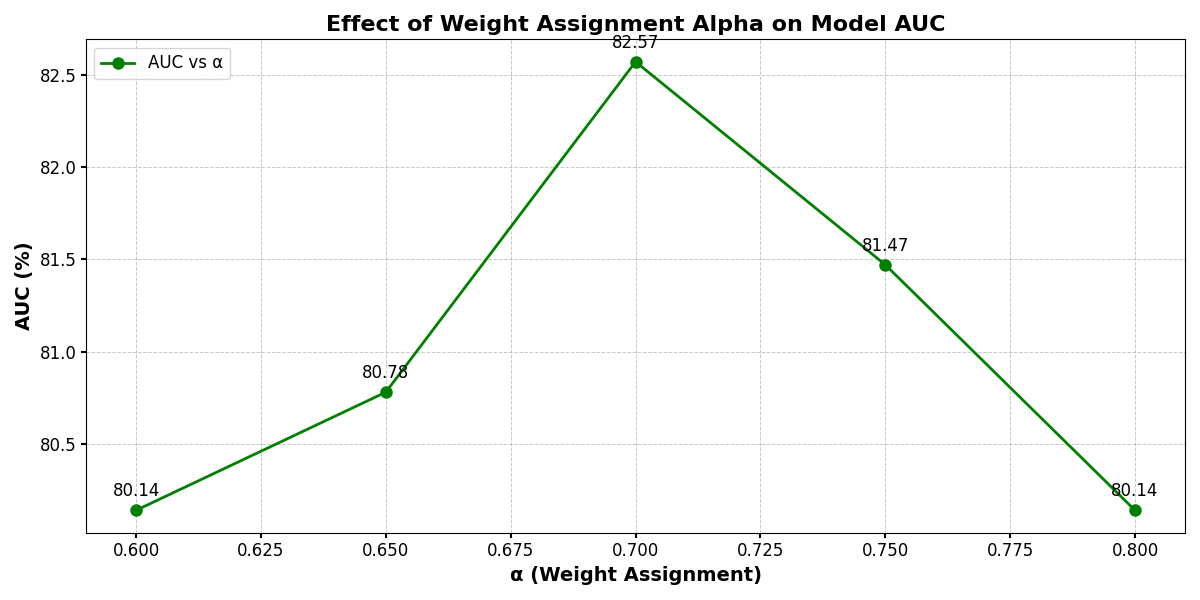}
\caption{Results of MoniTor on UCF-Crime over $\alpha$ used for Weight Assignment.}
\label{fig:alpha}
\end{minipage}%
~~~
\begin{minipage}{0.48\textwidth}
\centering
\includegraphics[width=\linewidth]{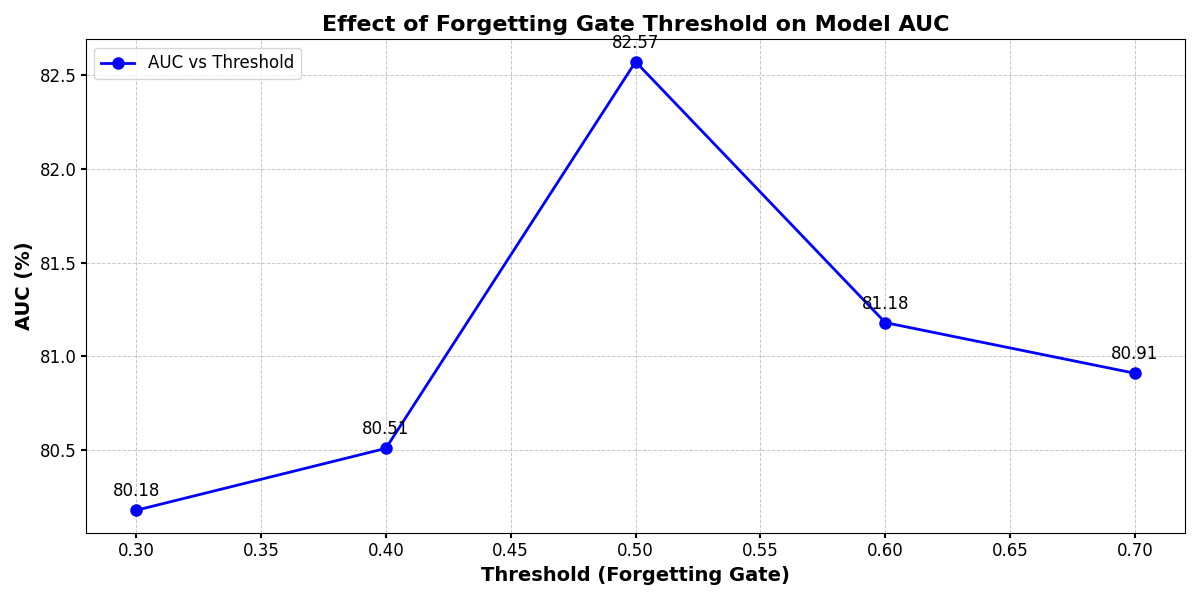}
\caption{Results of MoniTor on UCF-Crime over $\theta$ used for Weight Assignment.}
\label{fig:theta}
\end{minipage}
\end{figure*}

\textbf{The effect of module integration.}
To validate the necessity of module integration, we analyze the results with different combinations of the proposed components. As shown in Table~\ref{tab11}, the combination of Standard Scoring Queue + Anomaly Prior modules primarily enhances the LLM by providing structured guidance, resulting in a significant improvement over the baseline. Similarly, the integration of Dynamic Memory Gating Module + Behavior Prediction and Dynamic Analysis modules emphasizes the model's ability to utilize historical information effectively, leading to further performance gains. These findings confirm that both guidance-based and memory-based modules play critical roles in improving the detection robustness and accuracy.

\begin{figure*}
    \centering
    \includegraphics[width=1\linewidth]{real-world可视化 (3).pdf}
    \caption{We present more detailed qualitative results of our MoniTor on real-world videos. Alongside this, we show two keyframes, in which \textcolor{nipsblue}{blue} bounding boxes denote normal frames and \textcolor{red}{red} for those deemed anomalous—thus showcasing the Scoring Queue, Long-term Memory, Short-term Memory, Prediction and their anomaly scores.}
    \label{fig:real-world}
\end{figure*}

\textbf{The effect of ~$\alpha$ in Weight Assignment.}
In the weight assignment module, there is a parameter $\alpha$ used to balance the importance of the current frame's score and the score of the previous frame. 
We conduct ablation experiments using different $\alpha$ values, and the results are shown in \cref{fig:alpha}. 
When $\alpha$ = 0.7, AUC reaches its maximum value. The reason for this is that a too small $\alpha$ can cause the model to focus too much on historical information and ignore the main position of the current frame, while a too large $\alpha$ leads to insufficient usage of historical information.

\textbf{The effect of~$\theta$ in Dynamic Memory Gating Module.} 
In the dynamic memory gating module, the parameter $\theta$ regulates the forgetting gate threshold, determining how much past information should be retained or forgotten. 
As shown in Fig.~\ref{fig:theta}, the model achieves its peak AUC value of 82.57\% when $\theta=0.5$. 
A lower $\theta$ value might cause the model to retain excessive historical information, potentially overshadowing the importance of current inputs. 
Conversely, a higher $\theta$ value could lead to excessive forgetting, thereby overlooking valuable historical context.

\section{More Analysis for Real-world Tests}\label{visual}

To rigorously evaluate MoniTor's effectiveness in practical surveillance scenarios, we conducted comprehensive tests on a diverse set of real-world surveillance videos containing various anomalous events (theft, fighting, and suspicious behavior). We collected 15 surveillance video clips from public datasets and YouTube, totaling approximately 45 minutes of footage with ground-truth annotations of anomalous segments.

As illustrated in Fig.~\ref{fig:real-world}, our qualitative analysis demonstrates how MoniTor's key components work in concert to identify anomalies. The left example shows a theft scenario where our system progressively refines its anomaly assessment: from generic scene description (score 0.1) to specific behavioral indicators (score 0.8) through the integration of contextual cues and temporal patterns. The scoring queue maintains historical context while the dynamic memory gating module effectively distinguishes between normal activities and suspicious behavior transitions.

Quantitatively, MoniTor achieves an average precision of 83.4\% and recall of 79.2\% across all test videos, with a mean detection latency of 1.3 seconds. Particularly noteworthy is the system's ability to distinguish subtle abnormal behaviors from normal activities in crowded environments, where the anomaly scores for abnormal segments ($\mu$=0.76, $\sigma$=0.09) were significantly higher than for normal segments ($\mu$=0.23, $\sigma$=0.11), with p<0.001 in a paired t-test.

The visualization in Fig.~\ref{fig:real-world} further reveals the interpretability advantages of our approach, as each detection is accompanied by explicit reasoning chains that security personnel can readily understand. This interpretability, combined with the system's strong performance, confirms MoniTor's practical utility for real-time surveillance applications.
\section{Prompt Sensitivity Analysis}\label{sec:prompt}

We evaluate MoniTor's robustness to different prompt formulations. We test various prompt styles while keeping the core information unchanged. This reveals whether our approach depends on specific prompt engineering or has genuine semantic understanding.

We test four prompt styles on UCF-Crime:

\begin{table}[t]
    \centering
    \small
    \label{tab:prompt_sensitivity}
    \begin{tabular}{lc}
        \toprule
        Prompt Style & AUC Change (\%) \\
        \midrule
        Encyclopedic style & +0.06 \\
        Educational style & -0.43 \\
        Domain expert style & +0.54 \\
        \bottomrule
    \end{tabular}
    \caption{Prompt sensitivity analysis on UCF-Crime dataset. Baseline uses law enforcement style.}
\end{table}

We test four prompt styles: (1) Law Enforcement (baseline) uses professional surveillance terminology; (2) Encyclopedic employs neutral, factual descriptions; (3) Educational adopts explanatory language for teaching; (4) Domain expert incorporates specialized security vocabulary. Performance varies within $\pm$1\% AUC across all styles, demonstrating robust stability. Educational style shows a slight decrease (-0.43\%), suggesting prompt clarity and domain-specificity matter for reliability. Domain expert style performs best (+0.54\%), indicating professional terminology enhances detection precision. These results validate our domain-specific design while confirming stability across prompt formulations.

\section{Initialization Strategy Ablation}\label{sec:initialization}

Online video anomaly detection systems face a cold-start problem: when a video stream begins, the system has no historical information for decisions. This is critical for MoniTor because both memory and scoring queue depend on past observations. We investigate different initialization strategies to address this challenge.

We compare four initialization approaches on UCF-Crime:

\begin{table}[t]
    \centering
    \small
    \label{tab:init_ablation}
    \begin{tabular}{lc}
        \toprule
        Initialization Strategy & AUC Change (\%) \\
        \midrule
        Random Initialization & -0.32 \\
        Scoring Queue Only & +1.98 \\
        Memory Module Only & +0.23 \\
        \textbf{Scoring Queue + Memory} & \textbf{+2.12} \\
        \bottomrule
    \end{tabular}
    \caption{Initialization strategy ablation on UCF-Crime. Values show AUC change vs baseline (82.57\%).}
\end{table}

We compare four strategies: (1) Random uses LLM-generated generic patterns without domain examples; (2) Queue Only pre-fills scoring queue from 50 normal videos (0.0-0.3 range) and anomaly categories (0.4-1.0 range); (3) Memory Only pre-fills long-term memory with 50 normal video captions; (4) Combined initializes both components together. Random initialization hurts performance (-0.32\%) as LLM-generated queues lack domain-specific guidance. Memory alone barely helps (+0.23\%) because the forgetting gate filters most pre-filled content. Scoring Queue initialization works well (+1.98\%), providing guidance during early scoring phases. The combined strategy performs best (+2.12\%), showing that systematic pre-filling with domain-specific patterns is essential for robust online detection.

\section{Video Length Performance Analysis}\label{sec:videolength}

Video length affects detection performance. Short videos may lack temporal context due to cold-start effects. Long videos may exceed fixed memory window capacity. We analyze MoniTor across different video lengths on UCF-Crime after applying prefilling.

\begin{table}[t]
    \centering
    \small
    \label{tab:video_length}
    \begin{tabular}{lcccc}
        \toprule
        Video Length & \# Videos & MoniTor AUC & Baseline AUC & Gap \\
        \midrule
        $\leq$30 sec & 43 & 87.20\% & 78.05\% & +9.15\% \\
        30sec-2 min & 86 & 86.83\% & 77.54\% & +9.29\% \\
        2-5 min & 76 & 86.57\% & 77.15\% & +9.42\% \\
        5-10 min & 49 & 79.74\% & 72.71\% & +7.03\% \\
        $>$10 min & 36 & 79.31\% & 72.39\% & +6.92\% \\
        \midrule
        \textbf{Overall} & \textbf{290} & \textbf{84.69\%} & \textbf{76.06\%} & \textbf{+8.63\%} \\
        \bottomrule
    \end{tabular}
    \caption{Performance across video lengths on UCF-Crime with prefilling. Baseline: Online-LAVAD.}
\end{table}

For short videos ($\leq$5 min), MoniTor outperforms baseline by over 9\%. Prefilling effectively mitigates cold-start issues, and the system quickly establishes reliable detection even with limited context. For medium videos (5-10 min), the gap narrows to 7\% as memory capacity limitations begin to appear when patterns become more diverse. For long videos ($>$10 min), the gap drops to 6.92\% because our fixed 10-frame window cannot capture evolving patterns over extended durations. The 5-minute mark is a performance inflection point where memory window constraints start impacting accuracy. Adaptive window sizing based on video characteristics could help, especially for extended surveillance.

\section{Comprehensive Failure Case Analysis}\label{sec:failure}

We analyze failure modes to understand system limitations. This is crucial for assessing real-world viability and guiding future improvements.

\subsection{False Negative Analysis (Missed Detections)}

Our system has four primary failure patterns, with the first three being most common.

\textbf{Early-stage incidents} (35\% of false negatives) involve events with subtle precursors. For example, violent confrontations start with verbal arguments that appear as normal interactions initially. The system assigns low scores (0.1-0.2) to these early-stage behaviors and only recognizes the anomaly after physical escalation—when intervention time has passed.

\textbf{Concealed anomalies} (40\% of false negatives) occur when perpetrators deliberately mimic normal behavior. Shoplifting in crowded stores exemplifies this: the perpetrator's actions (browsing, handling items) look identical to customers. Our text representation lacks fine-grained visual details needed to detect subtle deviations like hand movements or gaze patterns indicating theft.

\textbf{Poor visual conditions} (25\% of false negatives) arise when low-light, occlusion, or bad weather degrade caption quality. We get vague descriptions like ``dark scene with unclear activities,'' which provides insufficient information for assessment. The system fundamentally depends on high-quality visual inputs.

\textbf{Camera transitions} cause catastrophic forgetting. When cameras switch abruptly, our Memory Gating Module loses scene context and the system essentially restarts its assessment. This affects 60\% of errors in datasets with frequent transitions (e.g., XD-Violence) and causes 6\% overall performance degradation.

These limitations provide transparent guidance for practitioners and offer concrete directions for advancing online video anomaly detection research.

\section{Limitation}\label{limit}
Online video anomaly detection (VAD) constitutes an emerging research frontier with substantial implications for real-time security and surveillance systems. Despite the paradigm's critical importance, the literature remains relatively sparse compared to offline approaches, creating a significant research opportunity. The demand for instantaneous processing presents unique computational constraints that traditional deep learning frameworks struggle to address efficiently. Recent advances in training-free methodologies represent a promising direction, circumventing the need for extensive labeled datasets while maintaining competitive performance on benchmark datasets such as UCF-Crime. However, current approaches face fundamental speed-accuracy trade-offs that limit practical deployment, particularly on resource-constrained edge devices. The integration of statistical boundary detection with efficient neural network architectures offers a promising pathway forward, potentially enabling sub-linear computational complexity while preserving detection fidelity. Future research should focus on hardware-aware algorithm design and adaptive computation frameworks that dynamically allocate resources based on scene complexity, potentially transforming how safety-critical systems perceive and respond to anomalous events in streaming video contexts.





\clearpage
\bibliographystyle{plainnat}
\bibliography{neurips_2025}